\documentclass[wcp]{jmlr}

\jmlrvolume{230}
\jmlryear{2024}
\jmlrworkshop{Conformal and Probabilistic Prediction with Applications}
\jmlrproceedings{PMLR}{Proceedings of Machine Learning Research}

\usepackage{amsmath}
\usepackage{amssymb}
\usepackage{graphicx}
\usepackage{url}
\usepackage{longtable}
\usepackage{array}
\usepackage{multirow, multicol}
\usepackage{booktabs}
\usepackage{float}
\usepackage{dsfont}
\usepackage{amsfonts}
\usepackage{algorithm}
\usepackage{algpseudocode}
\usepackage{bbm}
\usepackage{enumitem}
\usepackage{pifont}
\usepackage{tikz}
\usepackage{mathtools}
\usepackage{bm}
\usepackage{subcaption}
\usepackage{wrapfig}
\usepackage{verbatim}

\newcommand{\eg}{\emph{e.g.}}
\newcommand{\ie}{\emph{i.e.}}

\newcommand{\X}{\mathcal{X}}
\newcommand{\Y}{\mathcal{Y}}
\newcommand{\D}{\mathcal{D}}
\newcommand{\I}{\mathcal{I}}
\newcommand{\R}{\mathbb{R}}

\newcommand{\equivto}{\stackrel{\scalebox{0.5}{$\triangle$}}{=}}
\newcommand{\checkyes}{\ding{51}}
\newcommand{\checkno}{\ding{55}}

\title{Conformal time series decomposition with component-wise exchangeability}

\author{
    \Name{Derck W. E. Prinzhorn}\thanks{Equal contributions. Correspondence to $<$derck.prinzhorn@student.uva.nl$>$.} \Email{derck.prinzhorn@student.uva.nl}
    \AND
    \Name{Thijmen Nijdam}$^{*}$ \Email{thijmen.nijdam@student.uva.nl}
    \AND
    \Name{Putri A. {van der Linden}} \Email{p.a.vanderlinden@uva.nl}
    \AND
    \Name{Alexander Timans} \Email{a.r.timans@uva.nl} \\ \addr University of Amsterdam, The Netherlands
}

\editor{Simone Vantini, Matteo Fontana, Aldo Solari, Henrik Boström and Lars Carlsson}

\begin{document}

\maketitle

\begin{abstract}
    Conformal prediction offers a practical framework for distribution-free uncertainty quantification, providing finite-sample coverage guarantees under relatively mild assumptions on data exchangeability. However, these assumptions cease to hold for time series due to their temporally correlated nature. In this work, we present a novel use of conformal prediction for time series forecasting that incorporates time series decomposition. This approach allows us to model different temporal components individually. By applying specific conformal algorithms to each component and then merging the obtained prediction intervals, we customize our methods to account for the different exchangeability regimes underlying each component. Our decomposition-based approach is thoroughly discussed and empirically evaluated on synthetic and real-world data. We find that the method provides promising results on well-structured time series, but can be limited by factors such as the decomposition step for more complex data.
\end{abstract}


\section{Introduction}
\label{sec:intro}

Time series forecasting is a central task that sees widespread occurrence in various sectors such as finance, business, energy, or weather \citep{sezer2020financial, deb2017review, lim2021time}, and decision-makers value the ability to quantify the uncertainty associated with a given forecast \citep{makridakis2016forecasting, padilla2021uncertain}. Conformal prediction (CP) has emerged as a popular method for such uncertainty quantification, providing distribution-free, finite-sample coverage guarantees under assumptions on \emph{data exchangeability} \citep{shafer2008tutorial, vovk2005algorithmic}. However, data modalities such as time series are generally non-exchangeable due to their sequentially correlated nature, rendering applications of CP challenging. Recent CP approaches catered to non-exchangeable data such as \cite{barber2023conformal}, \cite{enbpi} or \cite{aci} relax these exchangeability assumptions through weighting strategies emphasizing sample contributions, or rely on alternate assumptions altogether. These methods, however, always operate directly on the original time series.

Alternatively, temporal signals can be represented through a set of components that reflect varying informative temporal patterns, obtained via a \emph{time series decomposition} (TSD, \cite{hyndman2018forecasting}). Such a decomposition provides a flexible way of capturing and analysing temporal correlations at different scales, and has been employed to improve forecasting quality or identify outliers \citep{theodosiou2011forecasting, wen2019robuststl}. Interestingly, the interplay between possible decomposition models and CP approaches for time series remains unexplored. 

\begin{figure}
    \centering
    \begin{minipage}{\textwidth}\includegraphics[scale=0.06]{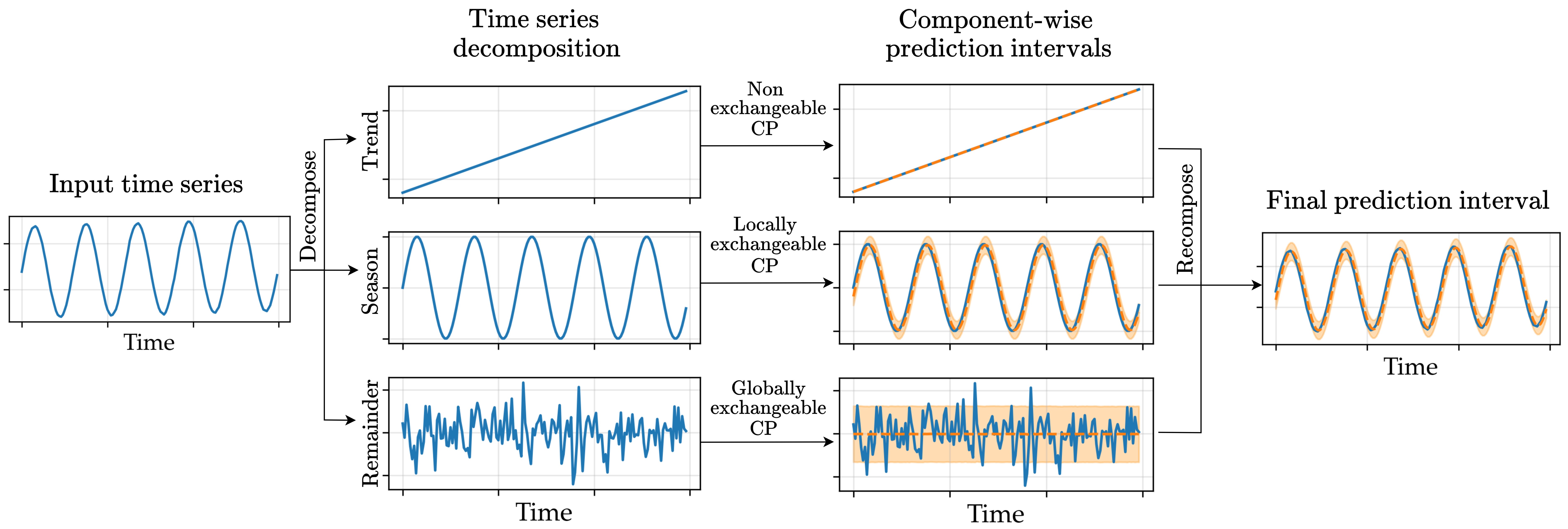}
    \caption{A high-level overview of our conformal time series decomposition approach. A time series signal is \emph{decomposed} into individual components, and each component is associated with a  specific regime of exchangeability (none, local or global). Relevant conformal methods are applied to each component, producing intermediate prediction intervals which are then \emph{recomposed} to obtain a prediction interval for the overall time series.}
    \label{fig:pipeline_decomp}
    \centering
    \end{minipage}
\end{figure}

In our work, we investigate this connection by leveraging TSD principles for conformal time series forecasting (see \autoref{fig:pipeline_decomp}). Our approach decomposes the time series and relates each extracted component to a specific underlying \emph{exchangeability regime}, which guides the application of tailored CP methods per component. Obtained component-wise prediction intervals (PIs) are then recomposed to obtain aggregated intervals for the overall time series. We validate our methods on both synthetic and real-world data, showing promising results when exchangeability assumptions seem satisfied. To the best of our knowledge, we are the first to thoroughly explore the intersection of conformal methods and time series decomposition. In summary, our key contributions include:
\begin{itemize}
    \item a proposed pipeline for conformal time series forecasting via time series decomposition, with several crucial design choices outlined and motivated; and
    \item conceptually relating temporal dependencies underlying each time series component to different notions of exchangeability, including novel weighting strategies for the seasonal time series component which are framed in terms of local exchangeability.
\end{itemize}

\section{Background: time series decomposition and exchangeability regimes}
\label{sec:background}

We begin by introducing notation for our time series and conformal prediction settings. We next provide general background on time series decomposition, and follow with details on selected CP methods for identified regimes of data exchangeability relevant to our approach.

\paragraph{Time series data.} We define a time series of length $T$ as a sequence of sampled covariate-response pairs $\D = \{(x_t, y_t)\}_{t=1}^{T} \subseteq \X \times \Y$ from $P_{XY}$. We denote feature and target spaces $\X \subseteq \R^d$ and $\Y \subseteq \R$, since we consider a univariate time series (as opposed to multivariate responses). Observed pairs are split into three disjoint sets: a training set $\D_{train}$, calibration set $\D_{cal} = \{(x_i, y_i)\}_{i=1}^{n}$ and test set $\D_{test} = \{(x_j, y_j)\}_{j=n+1}^{n_t
}$. $\D_{train}$ is used to fit a prediction model $\hat{f}_{\theta}: \X \rightarrow \Y$, $\D_{cal}$ is leveraged for any conformal procedure, and we evaluate on $\D_{test}$.

\paragraph{Conformal prediction.} In order to build coverage-controlling prediction sets, we first define a scoring function $s:\X \times \Y \rightarrow \R$ applied to $\D_{cal}$, resulting in a set of nonconformity scores $S = \{s(\hat{f}_{\theta}(X_i),Y_i)\}_{i=1}^{n} \equivto \{s_i\}_{i=1}^{n}$\footnote{Random variables are denoted upper-case ($X$) and their observed realizations as lower-case ($x$).}. These encode a desired notion of dissimilarity (\ie, nonconformity) between predictions and responses, such as the plain residual $|\hat{f}_{\theta}(X_i) - Y_i|$. The scores are leveraged to compute a conformal quantile $\hat{Q}(1-\alpha; \hat{F}(S))$, where $\hat{F}(S)$ denotes the empirical c.d.f. of the score set $S\,\cup\,\{+\infty\}$\footnote{The inclusion of $\{+\infty\}$ is a notational formality, see, \eg, \cite{barber2023conformal}.}. The target coverage level $(1-\alpha)$ is defined for a tolerated miscoverage rate $\alpha \in (0,1)$. For a new test sample $(X_{n+1},Y_{n+1})$, a valid prediction set for $X_{n+1}$ is then constructed as $\hat{C}(X_{n+1})=\{y \in \Y: \, s(\hat{f}(X_{n+1}), y) \le \hat{Q}(1-\alpha; \hat{F}(S))\}$. Assuming data exchangeability holds, the set $\hat{C}(X_{n+1})$ then guarantees the desired marginal coverage, \ie, we have that $\mathbb{P}_{\D_{cal}, \D_{test} \sim P_{XY}}(Y_{n+1} \in \hat{C}(X_{n+1})) \ge 1-\alpha$. For our regression setting, the sets $\hat{C}(X_{n+1}) \subseteq \R$ take the form of prediction intervals (PIs).

\subsection{Time series decomposition}
\label{subsec:ts-decomp} 

Time series decomposition (TSD) involves the analysis of a time series by decomposing it into several components, each representing distinct time-dependent patterns across varying durations. Methods for TSD can vary in the type of time-dependent patterns they extract (\eg, components in target or frequency domains), and how these are combined (\eg, explicitly or implicitly). Most commonly, one assumes some form of explicit decomposition model with components in target space, \ie, that of the observed responses $\{y_t\}_{t=1}^T$ (see, \eg, \cite{hyndman2018forecasting}). These components typically include a long-term \emph{trend}, reflecting the overarching direction of the series over extended periods; a \emph{seasonal} component, capturing recurrent patterns within a specific cycle, such as yearly or quarterly variations; and a noise term or \emph{remainder} component, accounting for fluctuations, irregularities and randomness not explained by the trend or seasonal influences.

Specifically, let $Y_t$ denote the random variable of the response at time $t$. Then, a TSD into trend $T$, seasonality $S$ and remainder component $R$ can be modelled as 
\begin{equation}
    Y_t = T_t \circ S_t \circ R_t,
    \label{eq:tsd}
\end{equation}
where $\circ$ is a binary operator, often chosen to be either addition ($+$) or multiplication ($\times$). Generally, an additive model is suitable when the time series is relatively stable and exhibits low volatility, while a multiplicative structure is useful when there is strong proportionality of components to the level of the time series, \ie, a dependency on the magnitude of the data \citep{hyndman2018forecasting}. Alternatively, data transformations such as into $\log$-space or via Box-Cox power transforms \citep{osborne2010improving} may alleviate such dependencies and permit the use of additive models nonetheless. 

\noindent Various methods exist for estimating $T$, $S$ and $R$. Classical decomposition approaches following \cite{macaulay1931smoothing} involve isolating the trend through a smoothing mechanism, such as a weighted moving average. The estimated trend can in turn be used to extract seasonal patterns by de-trending the time series and averaging over seasonal instances, where the cycle frequency (\eg, yearly, monthly or daily) is pre-specified. Finally, the remainder is defined as the portion of the signal that is not explained by either component. This general framework has been extended and improved upon by a range of TSD methods which vary in expressivity, assumptions and use-cases (see \autoref{sec:related}).

A popular approach is \emph{Seasonal-Trend decomposition based on LOESS} (STL) introduced by \cite{STLdecompositionCleveland}, which leverages local weighted polynomial regression for the decomposition (see \cite{cleveland1988locally} for details). STL continues to see use in recent applications due to favourable properties such as its non-parametric formulation, ability to handle varying cycle frequencies of unconstrained form, and outlier robustness due to its weighting mechanism \citep{hyndman2018forecasting, theodosiou2011forecasting}. It is thus also our employed decomposition method throughout the experiments (\autoref{sec:exp}).

Regardless of a particular choice, TSD methods remain constrained by the inherent limitations of an attempted decomposition, which assumes clear underlying patterns and is susceptible to real-world data irregularities, adversely affecting any decomposition results. Such irregularities can include data outliers, fluctuating seasonal patterns, complex systematic interactions (\eg, irregular trend behaviour), or inherently low signal-to-noise ratios \citep{wen2019robuststl}. While expectations on a proper decomposition for `well-behaved' time series may be warranted, \eg, by arguments such as Wold's decomposition \citep{miamee1988wold}, retrieving interpretable components remains challenging in practice.

\subsection{Conformal methods for different exchangeability regimes}
\label{subsec:exch-regimes}

Conformal prediction methods were first proposed assuming global exchangeability, \ie, data exchangeability across all samples \citep{vovk2005algorithmic}. We define data exchangeability similar to \cite{shafer2008tutorial} as follows:
\begin{definition}[Exchangeability]
    A sequence of random variables $X_1,\dots,X_n$ is exchangeable if for any permutation $\pi: \{1,\dots,n\} \rightarrow \{1,\dots,n\}$ with $n \geq 1$ we have that $$P(X_{1},\dots,X_{n}) = P(X_{\pi(1)},\dots,X_{\pi(n)}).$$
\label{thm:exch}
\end{definition}
\vspace{-10mm}
That is, the joint distribution of $X_1,\dots,X_n$ remains invariant to any permutation of indices. Notably, the setting of \emph{i.i.d} random variables is a special case thereof. Relaxing this assumption to weaker notions of exchangeability has proven fruitful and permitted broadening the application of CP methods to settings with (temporally) dependent data (see \autoref{sec:related}), for which \autoref{thm:exch} is violated.

Relevant to our problem setting, we identify three distinct notions of exchangeability of varying assumption strength: \emph{(i)} the strongest condition of global exchangeability; \emph{(ii)} a total lack of exchangeability, instead relying on alternate assumptions; and \emph{(iii)} in-between forms of local exchangeability based on weighted CP. We next describe selected CP methods for each of these three exchangeability regimes in more detail, which are subsequently leveraged for our TSD approach in \autoref{sec:method}.

\paragraph{Global exchangeability.} The well-known standard split CP \citep{papadopoulos2007conformal} belongs to this case. Herein, the set of nonconformity scores $S$ is obtained from a single-split calibration set to compute the conformal quantile $\hat{Q}(1-\alpha; \hat{F}(S))$. To alleviate dependency on the single split and improve adaptivity, principles from cross-validation can be employed \citep{kim2020jackknifeab, vovk2015cross}. Specifically, \cite{barber2020jackknife} propose the jackknife+ as a per-sample, and CV+ as a per-fold \emph{leave-one-out} approach. The joint set $\D_{train} \cup \D_{cal}$ is split into $K$ distinct folds and predictors $\hat{f}_{\theta, -k}$ are fitted with the $k$-th fold removed, which in turn acts as an unbiased set to compute out-of-fold nonconformity scores. For any new test sample $(X_{n+1}, Y_{n+1})$, both split and cross-conformal approaches assume data exchangeability of $\mathcal{D}_{cal} \cup \{(X_{n+1}, Y_{n+1})\}$ for coverage guarantees to hold. 

\paragraph{Lack of exchangeability.} In the presence of temporal dependency between samples, proposed CP methods employ update strategies to achieve target coverage over time under alternate assumptions (see \autoref{subsec:trend}), and provided guarantees do not require data exchangeability. Ensemble batch prediction intervals (EnbPI, \cite{enbpi}), a popular method, makes use of repeated updates to the calibration set. An ensemble of bootstrapped estimators is used to obtain a score set $S_t$ of \emph{leave-one-out} nonconformity scores, up to some timestep $t$. A prediction interval $\hat{C}(X_{t+1})$ for $t+1$ then uses a time-dependent quantile $\hat{Q}_{t+1}(1-\alpha; \hat{F}(S_t))$, where $S_t$ is updated in a rolling fashion for the next step as $S_{t+1} := (S_{t}\backslash \{s_1\}) \cup \{s_{t+1}\}$. Adaptive conformal inference (ACI, \cite{aci}), another popular approach, proposes quantile updating on the basis of previously observed coverage. That is, given the current timestep $t$, PIs for $t+1$ utilize a time-dependent quantile $\hat{Q}_{t+1}(1-\alpha_{t+1}; \hat{F}(S_t))$, where the target coverage level $\alpha_{t+1}$ is affected by (mis-)coverage at the previous step via the update $\alpha_{t+1} = \alpha_t + \gamma \, (\alpha - \mathbbm{1}[Y_t \notin \hat{C}(X_t)])$, with step size $\gamma > 0$. 

\paragraph{Local exchangeability.} A number of CP approaches operate in-between the two mentioned regimes. These primarily leverage weighting mechanisms to assign varying measures of importance to individual scores, where such importance may relate to sample distances in time, feature space, distribution, and others (see \autoref{sec:related}). A general formulation of \emph{weighted CP} is given by computing a conformal quantile $\hat{Q}(1-\alpha; \hat{F}_{w}(S))$, where $\hat{F}_{w}(S)$ is now the empirical c.d.f. over a weighted score set and takes the form
\begin{equation}
\hat{F}_{w}(S) = \sum_{i=1}^{n} \tilde{w}_i \, \delta_{s_i} + \tilde{w}_{n+1} \, \delta_{+\infty},
\label{eq:weight-cp}
\end{equation}
with $\delta_{s_i}$ denoting the dirac delta centered at score $s_i$, and $\tilde{w}_i$ its associated normalized weight such that $\sum_{i=1}^{n} \tilde{w}_i = 1$. \cite{barber2023conformal} suggest fixed weights which aim to minimize the total variational distance, \eg, by upweighting more recent samples in case of distribution shifts; whereas \cite{guan2023localized} employ data-dependent weights determined by feature distance via a `localizer' function, such as the kernel distance $\exp\{-h\,|X_i - X_{n+1}|\}$. 

Interestingly, a relationship between particular weighting schemes and exchangeability assumptions can be established. For instance, it is straightforward to see that \autoref{eq:weight-cp} with the choice $\tilde{w}_i = 1/(n+1) \,\,\forall i=1, \dots, n$ recovers standard split CP, and thus subsumes global exchangeability for coverage guarantees to hold. We hypothesize that similar relations can be established for locally restricted (approximate) data exchangeability, as illustrated in \autoref{subsec:season}. We suggest defining such local exchangeability similarly to \autoref{thm:exch} as below:
\begin{definition}[Local exchangeability]
    Consider a sequence of random variables $X_1,\dots,X_n$ and denote its index set by $\I := \{1, \dots, n\}$. For any subset $\I' := \{1, \dots, k\} \subset \I$ the associated random variables $X_1,\dots,X_{k}$ are locally (approximately) exchangeable if for any permutation $\tilde{\pi}: \I' \rightarrow \I'$ with $|\I'| \geq 1$ we have that $P(X_{1},\dots,X_{k}) \simeq P(X_{\tilde{\pi}(1)},\dots,X_{\tilde{\pi}(k)}).$
\label{thm:loc-exch}
\end{definition}

\section{Leveraging exchangeability regimes for time series components}
\label{sec:method}

We now explicitly match CP methods for varying regimes of exchangeability to time series components obtained from its decomposition. Our key insight is the correspondence between obtained components and their potential to fulfill respective exchangeability assumptions, permitting the application of tailored CP algorithms to each component. While the correspondence is straightforward for remainder and trend terms, we explore different weighting mechanisms for the seasonal term, and relate them to notions of local exchangeability by \autoref{thm:loc-exch}. Our exchangeability hypotheses are empirically challenged in \autoref{sec:exp}.

\subsection{Remainder component: globally exchangeable}
\label{subsec:remainder}

The remainder component $R$ (\autoref{eq:tsd}) accounts for fluctuations that are left unexplained by other components. Also referred to as error or noise $\epsilon$, this term is frequently paired with a systematic component $f_{\theta}(X)$ in many statistical models, typically via the additive structure $f_{\theta}(X) + \epsilon$. Additional assumptions may be placed on $\epsilon$, such as on \emph{i.i.d} sample draws or its distribution origin (\eg, the common assumption $\epsilon \sim N(0, \sigma^2)$). These conditions apply to a broad spectrum of (non-)parametric models, including linear regression, generalized linear models, regression splines, ARIMA, and more (see, \eg, \cite{hastie2009elements}). Similarly, STL and its regression mechanism are based on a linear additive structure, but do not impose any distributional form \citep{STLdecompositionCleveland}.

Aligned with these common assumptions, we hypothesize $R$ to satisfy a lenient \emph{i.i.d} condition. If the signal is perfectly decomposed according to \autoref{eq:tsd}, the remainder term will ideally reflect uncorrelated functional randomness only -- what is referred to as `white noise' in time series modelling \citep{parzen1966time}. While practicly unattainable due to approximation errors, one can nonetheless expect an accurate TSD to capture a large part of the non-exchangeable time series signal, thereby justifying a relaxed \emph{i.i.d} assumption on $R$ via global exchangeability. This permits the use of CP methods requiring such a condition, including standard split CP and its cross-validation variants. Specifically, we consider the CV+ algorithm by \cite{barber2020jackknife}, which minimizes dependency on a single calibration split and is observed to yield smaller prediction intervals than split CP. On the other hand, its finite-sample, distribution-free coverage guarantee is limited to $(1-2\alpha)$, similar to other cross-conformal approaches \citep{vovk2015cross}. In practice, empirical coverage tends to satisfy $(1-\alpha)$ and above, as noted by \cite{barber2020jackknife} and corroborated in our experiments.  

\subsection{Trend component: non-exchangeable}
\label{subsec:trend}

The trend reflects long-term temporal patterns in the time series, and captures the strongest proportion of temporally correlated signal in the data. As such, observed trend values $\{T_t\}_{t=1}^T$ exhibit strong serial correlation, and any permutation risks modifying their joint distribution, thus violating \autoref{thm:exch}. In such settings, one may directly leverage any proposed CP methods designed for non-exchangeable time series, relying on the relationship between a time series and its extracted trend. This strong correspondence is evident, for example, in the common practice of detrending, a preprocessing step used to achieve data stationarity \citep{wu2007trend}.

In our experiments, we rely on the two popular CP approaches EnbPI \citep{enbpi} and ACI \citep{aci}, which we briefly outlined in \autoref{subsec:exch-regimes}. Since data exchangeability is considered violated, these methods rely on alternate assumptions to derive nominal coverage guarantees. EnbPI assumes a linear model structure $Y_t = f_{\theta}(X_t)\,+\,\epsilon_t$, and a stationary and strongly mixing error process $\{\epsilon_t\}_{t=1}^T$\footnote{The \emph{i.i.d} setting can be highlighted as a special mixing case.}. Under such conditions, a lower bound on the procedure's under-coverage is derived in finite samples (\cite{enbpi}, Thm. 1), and for $t \rightarrow \infty$ asymptotically attains marginal coverage $(1-\alpha)$. ACI does not require additional assumptions beyond conditions on the step size $\gamma$ to provide bounds on the miscoverage rate (\cite{aci}, Prop. 4.1), which likewise recovers asymptotic marginal coverage at level $(1-\alpha)$. In both instances, we generally find that empirical coverage is achieved in finite samples. We refer to their respective works for further details.

\subsection{Seasonal component: locally exchangeable}
\label{subsec:season}

\begin{wrapfigure}[11]{r}{0.4\textwidth}
  \vspace{-1.8\baselineskip}
  \begin{center}
  \includegraphics[width=0.9\linewidth]{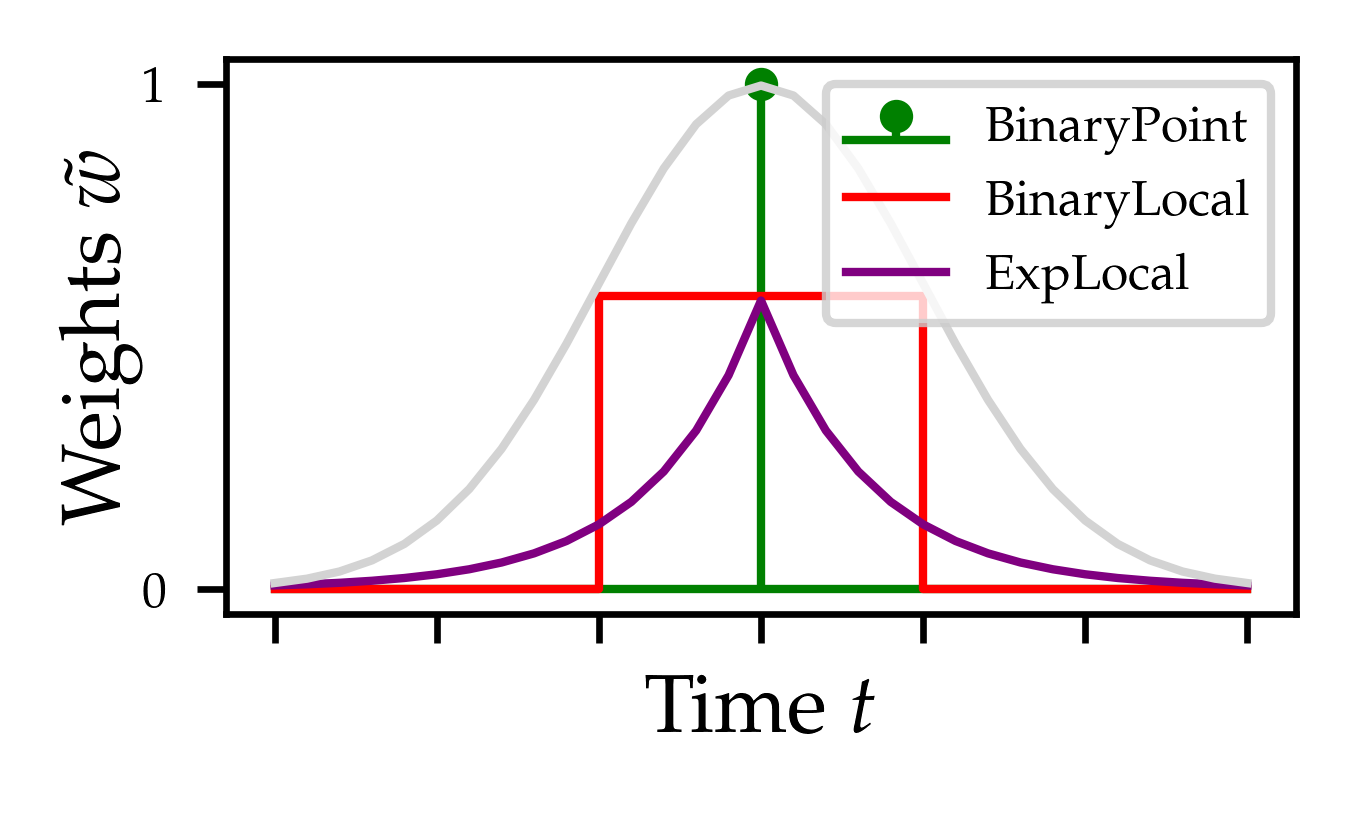}
  \end{center}
  \captionsetup{width=0.9\linewidth}
  \vspace{-1.8\baselineskip}
  \caption{Illustration of weighting schemes for an exemplary season (grey) with period $\tau=7$. The test sample's location is at the period peak, thus $\tau_{n+1} = 4$.}
  \label{fig:weights}
\end{wrapfigure}

The seasonal component $S$ is designed to capture regular patterns in time series, specifically periodically recurring variations like weekly or daily peaks. Although extracted patterns are time-dependent, their repetitive nature implies that such dependency might be confined to shorter time intervals. Specifically, for a consistent periodic schedule, strong temporal correlation is confined to samples within a single period (\eg, from one trough to the next), leading us to hypothesize that samples \emph{across} periods can be considered locally (approximately) exchangeable. For example, permuting the indices of samples at the period peaks or troughs is likely to have minimal impact on the overall pattern. In practice, we may test such local exchangeability by means of weighted CP according to \autoref{eq:weight-cp}, with particular weights $\tilde{w}$ reflecting different exchangeability conditions. We follow by detailing three such weighting schemes, qualitatively illustrated in \autoref{fig:weights}.

\paragraph{Weighting strategies.} Consider a seasonal pattern with fixed period length $\tau > 0$. In addition, let us denote by $\tau_{n+1}$ the position of a given test sample with respect to its period start. For example, if we consider the sample to be located at the season peak in \autoref{fig:weights}, then $\tau_{n+1} = 4$ for the given period length $\tau = 7$ (equating the particular time step). Next, let $\I_{cal} := \{1, \dots, n\}$ denote the index set of $\D_{cal}$. Our first approach \textbf{BinaryPoint} subsets $\I_{cal}$ based on the test sample's precise period location to obtain a binary weighting mechanism. That is, the filtered index set $\I_{cal}^{\text{BP}} \subset \I_{cal}$ takes the form
\begin{equation}
    \I_{cal}^{\text{BP}} := \{ i \in \I_{cal}: i \text{ mod } \tau = \tau_{n+1} \}.
\label{eq:idx-cal-bp}
\end{equation}
The employed weighting mechanism for \autoref{eq:weight-cp} is then given by $\tilde{w}_i = 1/(|\mathcal{I}_{cal}^{\text{BP}}| + 1)\cdot\mathbbm{1}[i \in \mathcal{I}_{cal}^{\text{BP}}]\,\,\forall i=1, \dots, n$, where equal weights are assigned to every sample whose index is in the filtered set, and zero otherwise\footnote{Observe that any split of $\D_{cal}$ can be defined as beginning at a seasonal period, which can take different shapes as long as the period length is consistent, \eg, trough to trough, peak to peak, \emph{etc.}}. Due to the risk of small calibration set sizes, we further consider an extension mechanism \textbf{BinaryLocal}, wherein BinaryPoint is expanded to include indices within local neighbourhoods $k>0$ of every sample in $\mathcal{I}_{cal}^{\text{BP}}$. In \autoref{fig:weights} such a neighbourhood is set to $k=1$, \ie, we include samples on either side of $\tau_{n+1}$ at the peak. As for BinaryPoint, the samples are weighted equally if included and zero otherwise, producing a hard binary weighting scheme. As an alternative, we also consider a softer weighting mechanism \textbf{ExpLocal}, which applies a repeated exponential decay to all samples in $\D_{cal}$. Specifically, an unnormalized weight for the $i$-th calibration sample is given by the decay $\lambda e^{-\lambda\cdot\Delta\tau_i}$, where $\lambda \in (0,1)$ denotes the decay rate and $\Delta\tau_i = |\tau_i - \tau_{n+1}|$ the sample's distance in regards to the target location $\tau_{n+1}$ for it associated period. Thus with increasing distance from $\tau_{n+1}$ the weights exponentially shrink, as visualised in \autoref{fig:weights}. 

Since our suggested weighting mechanisms are based on sample distance across time and assign fixed weights, they primarily relate to schemes discussed in \cite{barber2023conformal}. Such weights leveraging temporal distance are particularly useful for time series, where auto-regressive effects are prevalent. Inspired by \cite{guan2023localized}, we additionally explore data-dependent weighting based on feature distances in \autoref{app:feat-dist}.

\paragraph{Relation to local exchangeability.} An intuitive connection emerges between the described weighting mechanisms and underlying assumptions about local exchangeability following \autoref{thm:loc-exch}. For BinaryPoint (\autoref{eq:idx-cal-bp}), a highly selective calibration subset is chosen, and identical weights assume global exchangeability for all included samples. This equates a strong assumption of local exchangeability confined to a precisely defined calibration set. The condition is relaxed for BinaryLocal, wherein a larger set size is traded for a more approximate notion of local exchangeability, especially as the neighborhood expands or when the period length $\tau$ is small. ExpLocal further loosens the assumption by not selectively filtering calibration samples but rather weighting them based on their distance within periods. The associated notion of locality depends on factors like period length and decay rate $\lambda$, with the assumption approaching global exchangeability as $\lambda$ tends towards zero.

\subsection{Prediction interval recomposition}
\label{subsec:recomp}

After applying selected CP methods to each component of the time series, the component-level prediction intervals (PIs) must be \emph{recomposed} to establish final interval boundaries for the appropriate region in target space $\Y$. The recomposition approach is influenced by the chosen decomposition structure detailed in \autoref{eq:tsd}, linking directly to the choice of TSD model. Since we opt for STL, which is based on a linear decomposition, we utilize an additive structure for recomposition as well. Specifically, the lower and upper interval bounds for our conformal PI $\hat{C}(X_{n+1}) = [\hat{L}(X_{n+1}),\,\hat{U}(X_{n+1})]$ for $Y_{n+1}$ are computed as
\begin{equation}
\begin{split}
    \hat{L}(X_{n+1}) = \hat{L}_{T}(X_{n+1}) \,+\, \hat{L}_{S}(X_{n+1}) \,+\, \hat{L}_{R}(X_{n+1}) \\ \hat{U}(X_{n+1}) = \hat{U}_{T}(X_{n+1}) \,+\, \hat{U}_{S}(X_{n+1}) \,+\, \hat{U}_{R}(X_{n+1}), 
\end{split}
\label{eq:recomp}
\end{equation}
where the right-hand side represents the respective component-level bounds for trend $T$, season $S$ and remainder $R$. In our experiments, this linear recomposition sometimes proved overly conservative, for example when interactions between components tend towards a multiplicative nature. To obtain optimally tight PIs, identifying an appropriate recomposition model is as crucial as an accurate decomposition. In addition, one might consider adjusting coverage levels or a re-weighting of individual components based on observed PI widths. For instance, relaxing coverage requirements for components with a small magnitude can reduce the interval size of the recomposed PI without notably affecting its coverage.

\paragraph{Aggregated coverage guarantee.} Different CP methods can be employed across time series components, which complicates providing a clear statement about the nominal coverage for the recomposed PI detailed in \autoref{eq:recomp}. Under simplified conditions, a loose lower bound on nominal coverage is obtainable at level $1 - 3\alpha$ (see \autoref{app:cov-pi}).  In practice, traditional CP methods offer finite-sample guarantees, contrasting with the asymptotic guarantees provided by time series approaches. Additionally, any coverage guarantees applicable to settings of local exchangeability are highly dependent on specific data conditions, which limits the feasibility of generalizations. For example, the coverage bounds cited in \cite{barber2023conformal} are significantly influenced by the total variational distance term, whereas \cite{guan2023localized} highlight the need for careful tuning of the miscoverage rate to avoid overcoverage in practice. Consequently, deriving an overall coverage guarantee for the recomposed PI, and its practical applicability, remain open questions. The effectiveness of such guarantees is likely to be substantially affected by modelling factors such as decomposition quality and chosen weighting schemes.

\section{Related work}
\label{sec:related}

\paragraph{Time series decomposition.} Building upon classical TSD concepts \citep{macaulay1931smoothing}, popular techniques include X-11 \citep{cleveland1976x11}, SEATS \citep{dagum2016seasonal}, TBATS \citep{livera2011forecasting} and STL \citep{STLdecompositionCleveland}. We refer to \cite{esling2012time} for an older survey and \cite{mbuli2020decomposition} for a more recent overview with applications to power systems. In particular, STL sees continued use in recent works due to its favourable properties over, \eg, X-11 or SEATS \citep{hyndman2018forecasting}. This includes domain-specific applications such as traffic, weather or air quality \citep{huo2019long, he2022weatherstl, li2023airstl}, as well as extensions tackling multi-seasonality \citep{bandara2021mstl, trull2022multiple}, robustness \citep{wen2019robuststl, wen2020fast}, online settings \citep{he2023oneshotstl, mishra2021onlinestl} or uncertainty \citep{krake2024uncertainty, dokumentov2022str}. However, the connection between TSD models such as STL and CP frameworks remains unexplored. TSD models considering alternative decomposition structures (\ie, not into trend, seasonality and remainder terms) include theta-differences \citep{assimakopoulos2000theta}, wavelet \citep{soltani2002use} and spectral models \citep{bonizzi2014singular}. More recently, implicit decomposition structures in deep learning-based time series models have been explored, \eg, via self-attention mechanisms in transformers \citep{jiang2022bridging, wang2022learning, zhou2022fedformer, wu2021autoformer, lin2021ssdnet}. However, retrieved structures are typically fuzzy and sensitive to model specification; see \cite{wen2022transformers} for a recent overview.

\paragraph{Conformal prediction for time series.} Recent CP approaches for time series typically employ updating strategies to account for time-dependency of observations. \cite{aci} fix the calibration set but dynamically adjust the conformal quantile on the basis of observed coverage. \cite{zaffran2022adaptive} build upon this idea with a more adaptive weighted quantile, while \cite{angelopoulos2023conformal} design a quantile tracker borrowing ideas from control theory. \cite{auer2023conformal} obtain quantile weights by leveraging the attention mechanism of a Hopfield Network model. Alternatively, \cite{enbpi} select a fixed quantile and instead periodically update the underlying calibration set. \cite{Jensen_2022} combine this approach with CQR \citep{cqr} for quantile regressors, while \cite{xu2023sequential} replace the fixed quantile with a quantile estimator for more flexibility. While all the above consider a single time series, CP methods have been also proposed for the multivariate case. Assuming exchangeability between the observed time series, these can be used to build individual calibration sets for multiple prediction steps. \cite{stankeviciute2021conformal} treat these steps as independent, \cite{sun2022copula} account for dependency with a copula-based quantile, and \cite{xu2024conformal} design ellipsoidal sets. \cite{lin2022conformal} suggest quantile updates that account for coverage across both time steps and time series. 

\paragraph{Local exchangeability.} CP methods controlling for global exchangeability violations (and not specifically designed for time series problems) rely on notions of locally restricted or \emph{weighted} exchangeability. In practice, this suggests subsetting or reweighting of calibration samples by some notion of importance. \cite{tibshirani2019conformal} suggest weights determined by the likelihood ratio of calibration and test samples to counter covariate shifts. Applications include \cite{lei2021conformal} for counterfactuals, \cite{fannjiang2022conformal} for biomolecular design, and \cite{candes2023conformalized} for survival analysis. \cite{podkopaev2021distribution} extend ideas to account for potential shifts in target space. \cite{guan2023localized} propose weights determined by kernel distances in feature space, which is leveraged by \cite{hore2023conformal} for improved test-conditional coverage. \cite{gyorfi2019nearest} suggest a nearest-neighbour reordering of scores based on feature distances, while \cite{amoukou2023adaptive} and \cite{han2022split} reweigh implicitly via density estimates of the empirical c.d.f. $\hat{F}(S)$. Recently, \cite{barber2023conformal} suggest fixed weights on the basis of total variational distances quantifying sequence exchangeability. Conceptually related to weighted CP are \emph{mondrian} CP methods, wherein $\D_{cal}$ is partitioned according to different requirements and exchangeability is assumed to hold within subgroups \citep{cauchois2021knowing, jung2022batch, toccaceli2019combination}.

\section{Experiments} 
\label{sec:exp}

We next empirically verify the effectiveness of our TSD approach across a range of time series datasets. In order to validate our design choices, we first evaluate on a synthetic time series with known decomposition structure. We then test our approach on three real-world time series from different domains: \emph{San Diego energy consumption}\footnote{\url{https://github.com/pratha19/Hourly_Energy_Consumption_Prediction}}, \emph{Rossman store sales}\footnote{\url{https://github.com/juniorcl/rossman-store-sales}} and \emph{Beijing air quality}\footnote{\url{https://github.com/Afkerian/Beijing-Multi-Site-Air-Quality-Data-Data-Set}} \citep{air-quality-data}. We begin by outlining our experimental setup, which includes employed regression models, CP baselines and evaluation metrics. Our code is publicly available at \url{https://github.com/dweprinz/CP-TSD}.

\paragraph{Model setup.} We consider three different regression models $\hat{f}_\theta$ for the underlying predictor: a simple linear model, an MLP regressor, and a gradient boosting model. We train the models in an auto-regressive fashion, \ie, model covariates at time $t$ may include previous observations $y_{t-k},\dots,y_{t-1}$ up to lag order $k$ alongside other available features. In order to handle extrapolation for non-stationary data, the gradient boosting regressor is trained on differenced data $\Delta y_t = y_t - y_{t-1}$. Such auto-regressive training procedures are in line with our \emph{sequential} setup, wherein we simulate receiving a single observation per time step in a streaming fashion\footnote{Our approaches are extendable to batch and online settings, which we do not consider here.}. Our regressors are implemented via \verb|scikit-learn| \citep{pedregosa2011scikit} and minimally tuned (see \autoref{app:rmse}), while employed conformal methods leverage the library \verb|MAPIE| \citep{taquet2022mapie} when possible.

Unless stated otherwise, our default target coverage level is fixed at $90 \%$ $(\alpha=0.1)$. Throughout, we employ STL \citep{STLdecompositionCleveland} for time series decomposition, as motivated in \autoref{subsec:ts-decomp}. Our TSD models are compared using the different weighting schemes described in \autoref{subsec:season} for the seasonal component, while CP algorithms EnbPI and CV+ are employed for the trend and remainder terms, respectively. Results on real-world data in \autoref{tab:real_samp} are reported for the linear regressor, which proved most robust in terms of hyperparameter selection. All results are averaged across multiple seeds. Further results across different combinations of methods and coverage levels can be found in \autoref{app:res} and \autoref{app:res-diff-alphas}. Predictive model performance as measured via RMSE is reported in \autoref{app:rmse}.

\paragraph{Baselines.} We baseline our TSD approaches against CP algorithms EnbPI and ACI in two ways. Firstly, we consider their direct application to the raw time series signal without a decomposition (marked as (\checkno) in results tables below). This baseline aims to quantify performance differences attributable to the employed TSD model, or an attempted decomposition in the more general sense. Additionally, we consider their application to each time series component \emph{after} applying a TSD model. These decomposition baselines help quantify the efficacy of tailoring CP methods to the exchangeability regimes underlying each component, as opposed to repeatedly applying the same (non-exchangeable) procedures.

\paragraph{Evaluation metrics.} We employ standard metrics to assess the two key desiderata of \emph{validity} and \emph{efficiency} for our conformal approaches. We measure validity via the \emph{prediction interval coverage probability} (PICP) -- also referred to as empirical coverage -- and efficiency via the \emph{prediction interval average width} (PIAW). These are defined as 
\begin{equation}
    \text{PICP} = \frac{1}{n_t}\sum_{j=1}^{n_t} \mathbbm{1}[Y_j \in \hat{C}(X_j)] \quad \text{ and } \quad \text{PIAW} = \frac{1}{n_t}\sum_{j=1}^{n_t} |\hat{C}(X_j)|,
    \label{eq:metrics}
\end{equation}
and metrics are computed equivalently for the overall time series (both in raw and recomposed forms) or specific components using respective prediction intervals\footnote{By annotating \autoref{eq:metrics} with component indices $T$, $S$ or $R$ respectively, which we omit for clarity.}. 

\begin{figure}[t]
    \centering
    \begin{minipage}[b]{0.49\linewidth}
        \includegraphics[width=\linewidth]{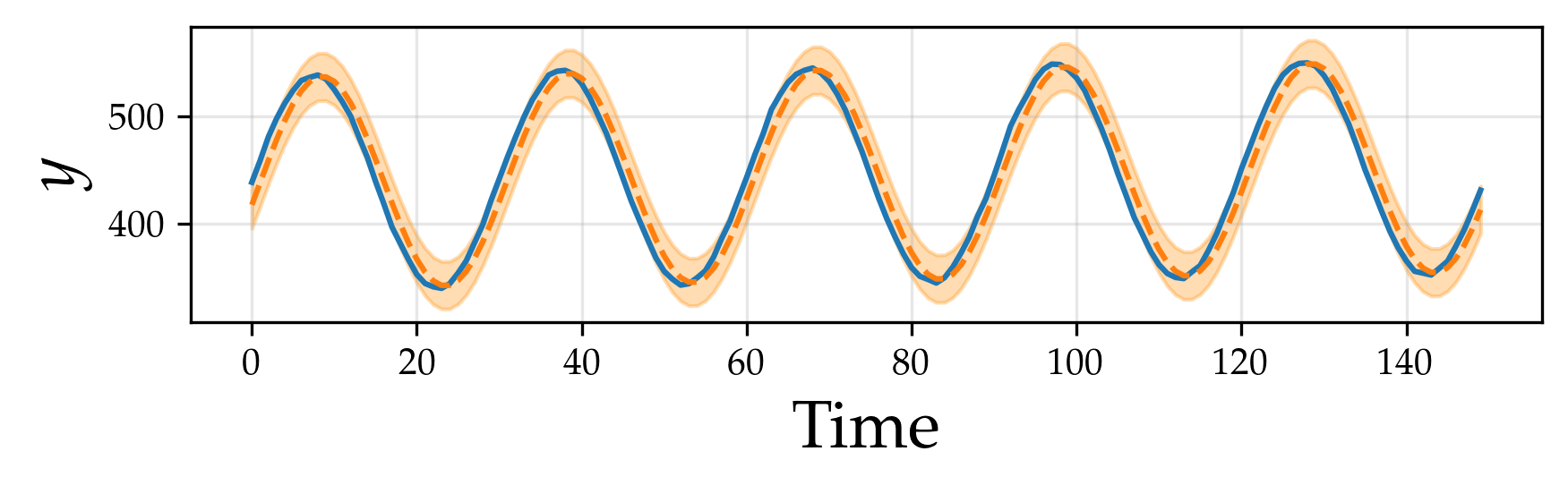}
    \end{minipage}
    \hfill
    \begin{minipage}[b]{0.49\linewidth}
        \includegraphics[width=\linewidth]{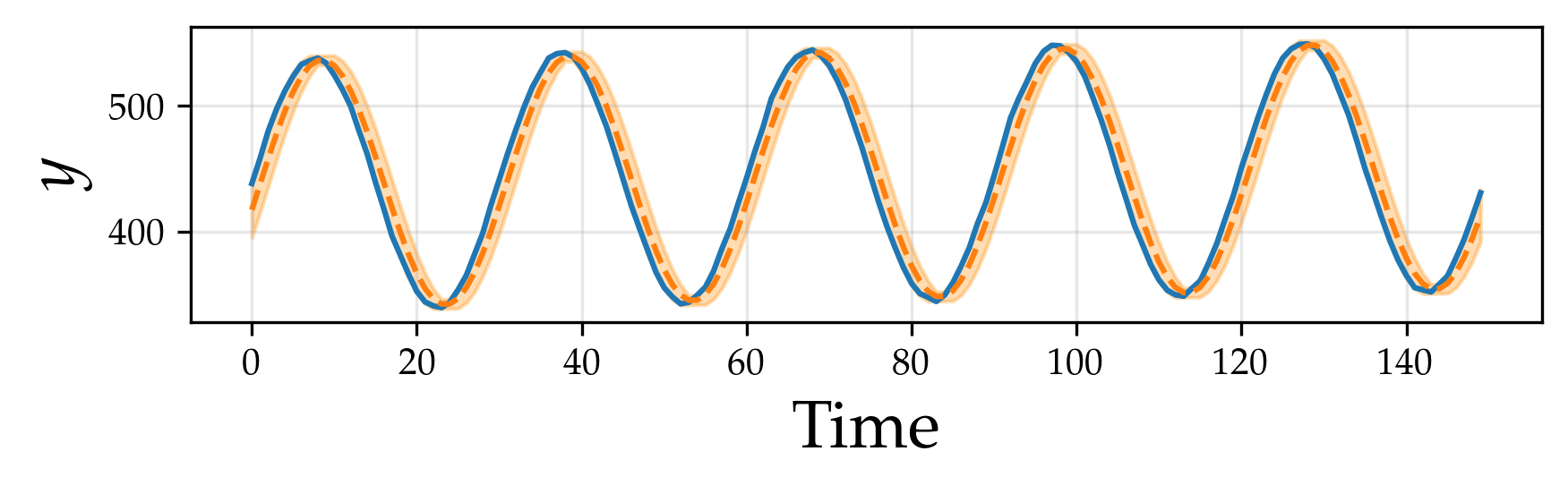}
    \end{minipage}
\caption{Qualitative results for a segment of the synthetic time series for the EnbPI decomposition baseline (\emph{left}) and our TSD approach with BinaryPoint (\emph{right}). We observe more efficient PIs across all time steps, but in particular at seasonal peaks and troughs.}
\label{fig:synth}
\vspace{-3mm}
\end{figure}

\begin{table}[b]
    \caption{Results on synthetic data for a fixed target coverage of $90 \%$ $(\alpha=0.1)$. Table entries for our TSD approaches denote different weighting schemes for the seasonal component, while EnbPI and CV+ are used for trend and remainder. Overall best results are \underline{underlined}, while best decomposition results are \textbf{boldened}.}
    \label{tab:synth_samp}
    \centering
    \scalebox{0.85}{
    \begin{tabular}{lccccccc}
     & & \multicolumn{2}{c}{\textbf{Linear Reg.}} & \multicolumn{2}{c}{\textbf{MLP Reg.}} & \multicolumn{2}{c}{\textbf{Gradient Boosting}} \\
    \cmidrule[\heavyrulewidth](lr){3-4}
    \cmidrule[\heavyrulewidth](lr){5-6}
    \cmidrule[\heavyrulewidth](lr){7-8}
    \multicolumn{1}{c}{\multirow{-1}{*}{\textbf{Method}}} & \textbf{Decomposed} & $PICP$ & $PIAW$ ($\downarrow$) & $PICP$ & $PIAW$ ($\downarrow$) & $PICP$ & $PIAW$ ($\downarrow$) \\
    
     \toprule
     EnbPI & \checkno &$0.889$ & $41.694$ & $0.887$ & $41.626$ & $0.897$ & $11.922$ \\
     EnbPI & \checkyes & $0.985$ & $44.013$ & $0.985$ & $44.197$ & $0.987$ & $12.536$ \\
     ACI & \checkno & $0.898$ & $42.065$ & $0.898$ & $42.168$ & $0.900$ & $11.991$ \\
     ACI & \checkyes & $0.985$ & $43.995$ & $0.986$ & $44.237$ & $0.994$ & $554.298$ \\
     \midrule
     BinaryPoint (Ours) & \checkyes & $0.946$ & \underline{\boldsymbol{$29.755$}} & $0.947$ & \underline{\boldsymbol{$29.771$}} & $0.947$ & \underline{\boldsymbol{$9.524$}} \\
     BinaryLocal (Ours) & \checkyes & $0.994$ & $41.323$ & $0.994$ & $41.347$ & $0.990$ & $11.942$ \\
     ExpLocal (Ours) & \checkyes &  $0.994$ & $44.086$ & $0.993$ & $44.009$ & $0.990$ & $12.510$ \\
     \bottomrule
    \end{tabular}
    }
    \vspace{-2mm}
\end{table}

\subsection{Synthetic data: time series with known decomposition structure}
\label{subsec:exp-synthetic}

We first test our model on a synthetic time series that follows a generative process with known decomposition structure. To this end, we construct a time series with an additive decomposition (in line with \autoref{eq:tsd}) consisting of the following components:
\begin{equation*}
    T_t = 0.1t, \quad S_t = 100 \cdot \sin \left(2\pi \cdot \frac{t}{30} \right), \quad R_t \sim \mathcal{N}(0, 1), \quad t=1,\dots,T.
\end{equation*}
\noindent In this controlled setting we identify the trend as non-exchangeable, the seasonal term as locally exchangeable, and the remainder as globally exchangeable, thus applying our TSD approaches leveraging component-specific CP methods. 

Results for different regressors are displayed in \autoref{tab:synth_samp}. Firstly, a comparison of baselines on the raw and decomposed time series suggest that a TSD approach is meaningful, since similarly sized PIs are obtained at much higher coverage. Given the observed overcoverage tendency, a tuning of nominal coverage levels would permit trading said coverage for improved interval widths, as suggested in \autoref{subsec:recomp}. Our TSD approaches with BinaryLocal and ExpLocal record similar behaviour, but achieve marginally tighter PIs than the decomposition baselines. Finally, BinaryPoint is able to fully exploit the sinusoidal pattern underlying the seasonal term, obtaining much-improved intervals over other TSD approaches and baselines. That is, the underlying assumption on strong local exchangeability is warranted by the data's consistent and accurately retrievable seasonality. This behaviour is highlighted in \autoref{fig:synth}, where the largest reductions in PI width are observed at the seasonal extrema.

\subsection{Real-world data}

\begin{figure}[t]
    \centering
    \includegraphics[width=1.0\textwidth]{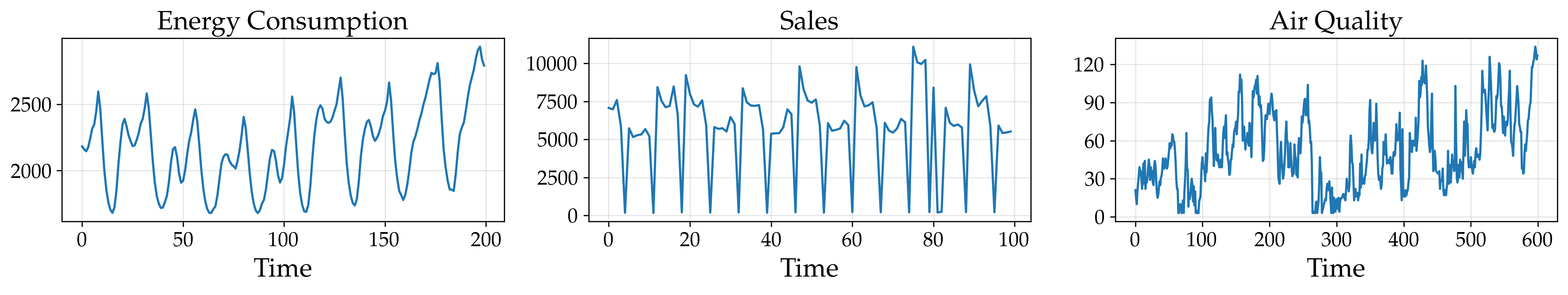}
    \includegraphics[width=1.0\textwidth]{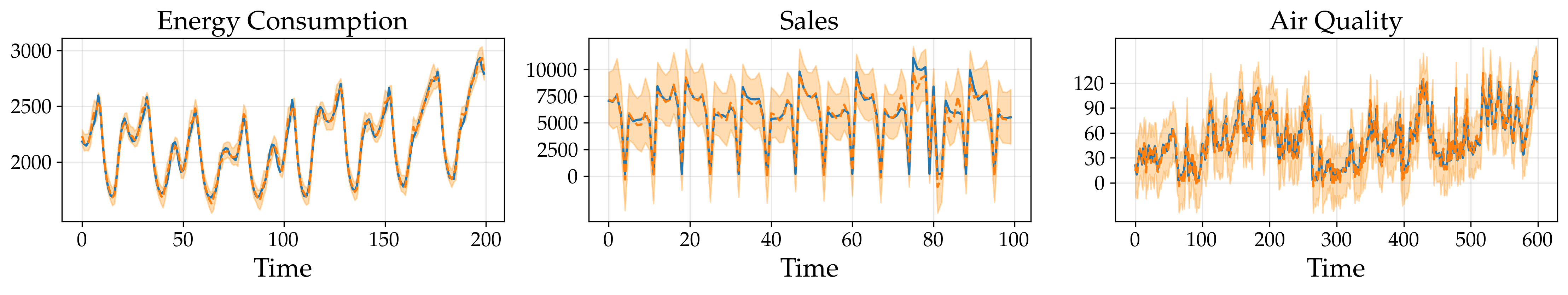}
    \caption{Segments of the three considered real-world time series from multiple domains. \emph{From left to right}: Different time series complexities are observed as we consider \emph{San Diego energy consumption} with a shifting trend, \emph{Rossman store sales} with varying seasonalities, and \emph{Beijing air quality} exhibiting stronger irregularity and noise. The second row includes prediction intervals leveraging our TSD approaches with the best-performing seasonal weights following \autoref{tab:real_samp} (\emph{left to right}: BinaryPoint, ExpLocal, ExpLocal).}
    \label{fig:real-world-datasets}
    \vspace{-4mm}
\end{figure}

\begin{table}[b]
    \centering
    \caption{
    Results on real-world data for a fixed target coverage of $90 \%$ $(\alpha=0.1)$ and linear regressor. Table entries for our TSD approaches denote different weighting schemes for the seasonal component, while EnbPI and CV+ are used for trend and remainder. Overall best results are \underline{underlined}, while best decomposition results are \textbf{boldened}.}
    \label{tab:real_samp}
    \scalebox{0.85}{
    \begin{tabular}{lccccccc}
    & & \multicolumn{2}{c}{\textbf{Energy}} & \multicolumn{2}{c}{\textbf{Sales}} & \multicolumn{2}{c}{\textbf{Air Quality}} \\
    \cmidrule[\heavyrulewidth](lr){3-4}
    \cmidrule[\heavyrulewidth](lr){5-6}
    \cmidrule[\heavyrulewidth](lr){7-8}
    \multicolumn{1}{c}{\multirow{-1}{*}{\textbf{Method}}} & \textbf{Decomposed} & $PICP$ & $PIAW$ ($\downarrow$) & $PICP$ & $PIAW$ ($\downarrow$) & $PICP$ & $PIAW$ ($\downarrow$) \\
     \toprule
     EnbPI & \checkno & 0.907 & 222.866 & 0.892 & 1572.343 & 0.902 & \underline{42.918} \\
     EnbPI & \checkyes & 0.964 & 243.005 & 1.0 & 5303.825 & 0.954 & 62.721 \\
     ACI & \checkno & 0.901 & \underline{218.301} & 0.895 & \underline{1539.925} & 0.901 & 43.549 \\
     ACI & \checkyes & 0.960 & \textbf{229.425} & 1.0 & 5447.390 & 0.958 & 63.445 \\
     \midrule
    BinaryPoint (Ours) & \checkyes  &  $0.971$ & $234.969$ &$1.0$ & $6045.373$ & $0.958$ & $67.014$\\
    BinaryLocal (Ours) & \checkyes&$0.967$ & $240.710$ &$1.0$ & $5426.439$&  $0.955$ & $62.716$ \\
    ExpLocal (Ours) & \checkyes &$0.967$ & $240.230$  & $1.0$ & \boldsymbol{$5248.193$} & $0.954$ & \boldsymbol{$62.240$} \\
     \bottomrule
    \end{tabular}
    }
    \vspace{-3mm}
\end{table}

We proceed by evaluating our approach on real-world datasets from different domains, of which exemplary segments are shown in \autoref{fig:real-world-datasets}. Such real-world time series are generally more volatile and contain less structured regularity. This poses challenges for any TSD model, in particular regarding retrieved seasonal patterns. While a relatively stable seasonality is observed for energy consumption, the sales data exhibits more erratic layered seasons, and the air quality dataset shows no clearly discernible patterns. Thus, the considered time series exhibit varying levels of complexity affecting both prediction and decomposition quality, as corroborated by empirical results. Dataset descriptions and detailed result tables for all three datasets can be found in \autoref{app:res}, while qualitative decomposition plots are in \autoref{app:decomp}.

\noindent Results for all three datasets are summarized in \autoref{tab:real_samp}. Notably, there is a discrepancy in obtained PI widths between baseline methods applied to the raw and decomposed time series, particularly evident for the sales data. This strongly relates to the practical limitations of an attempted decomposition, in that inaccurately retrieved components propagate down-stream and affect the performance of subsequently applied CP algorithms (see also \autoref{subsec:ts-decomp}). For example, STL is only capable of retrieving a single seasonal pattern, whereas the sales data might exhibit multiple seasonalities at different frequencies. This limitation also affects the performance of our TSD approaches, preventing them from surpassing time series-specific CP methods used on the raw signals directly. Among decomposition-based approaches, however, leveraging exchangeability regimes does appear to enhance performance for more complex datasets, with ExpLocal achieving best results on sales and air quality data. In contrast to synthetic data, the irregular seasonal patterns in real-world data benefit from softer weighting mechanisms like BinaryLocal and ExpLocal, promoting associated notions of approximate local exchangeability (\autoref{thm:loc-exch}). Overcoverage tendencies caused by our recomposition model in \autoref{eq:recomp} could be addressed by dataset-specific tuning, \eg, on nominal coverage levels. Particularly for more consistent time series such as energy consumption, such adjustments may help narrow the performance gaps observed.

\subsection{Comparison of methods for specific time series components}

\begin{figure}[t]
    \centering
    \includegraphics[width=0.95\textwidth]{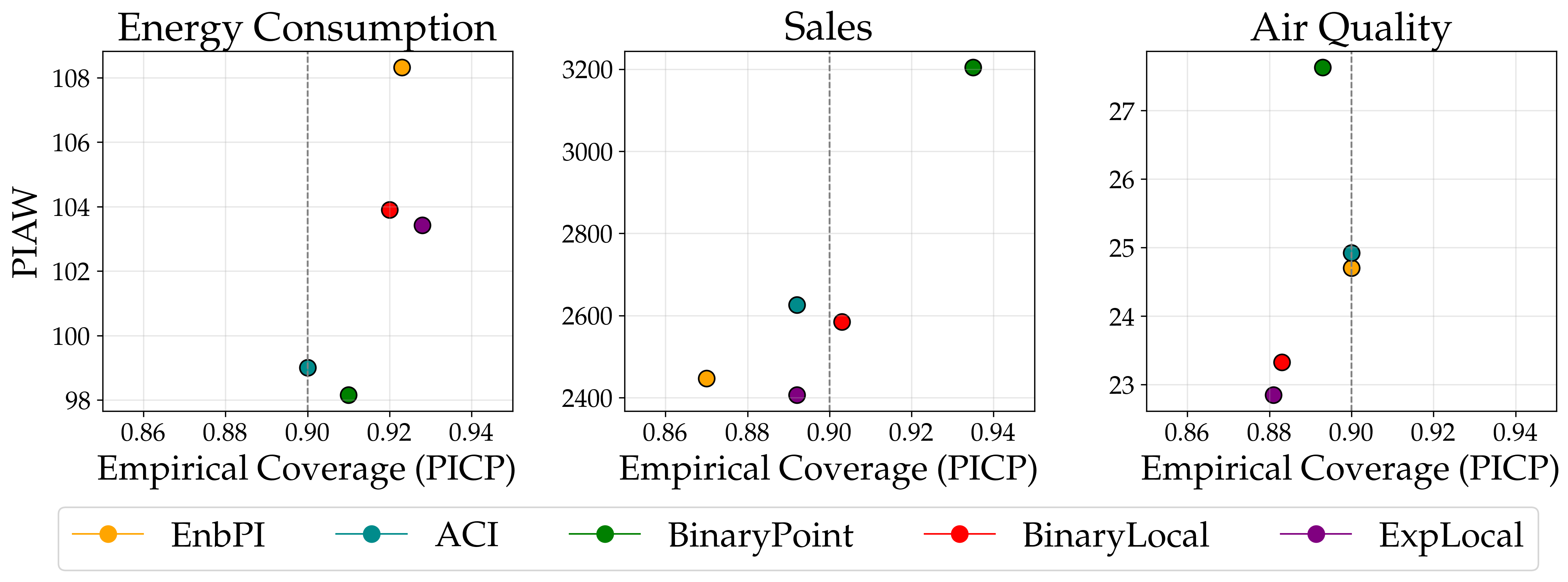}
    \caption{A comparison of decomposition baselines EnbPI and ACI and our weighting mechanisms (\autoref{subsec:season}) across real-world datasets. Metrics are computed based on obtained intervals for the \emph{seasonal} component only. The dashed line marks target coverage of $90\%$.}
    \label{fig:season-res}
\end{figure}

We now more closely examine and compare parts of our TSD approaches to decomposition baselines for individual components, particularly focusing on the seasonal and remainder terms, which are more susceptible to errors (see \autoref{app:rmse}). This analysis allows us to better understand how assumptions on underlying exchangeability impact the effectiveness of CP methods we employ on real-world data.

\paragraph{Seasonal component.} In \autoref{fig:season-res}, we evaluate our weighting mechanisms -- BinaryPoint, BinaryLocal, and ExpLocal -- against non-exchangeable baselines ACI and EnbPI, assessing the suitability of local exchangeability concepts in handling irregular real-world seasonal patterns. BinaryPoint performs best on energy consumption data, but yields significantly wider PIs for the other datasets, where BinaryLocal and particularly ExpLocal show strong performance. Conversely, EnbPI and ACI generally produce slightly broader PIs but maintain more consistency across datasets. These findings indicate that the effectiveness of employed CP methods is indeed highly dependent on a dataset's ability to meet its exchangeability assumption. Stronger exchangeability assumptions may benefit datasets with consistent seasonal patterns, while more approximate notions better suit complex and variable seasonality. However, it remains unclear whether using locally exchangeable CP methods for seasonal components offers a distinct advantage over generic time series CP methods like EnbPI, which are less dependent on satisfying such specific assumptions. 

\begin{figure}[t]
    \centering
    \includegraphics[width=0.95\textwidth]{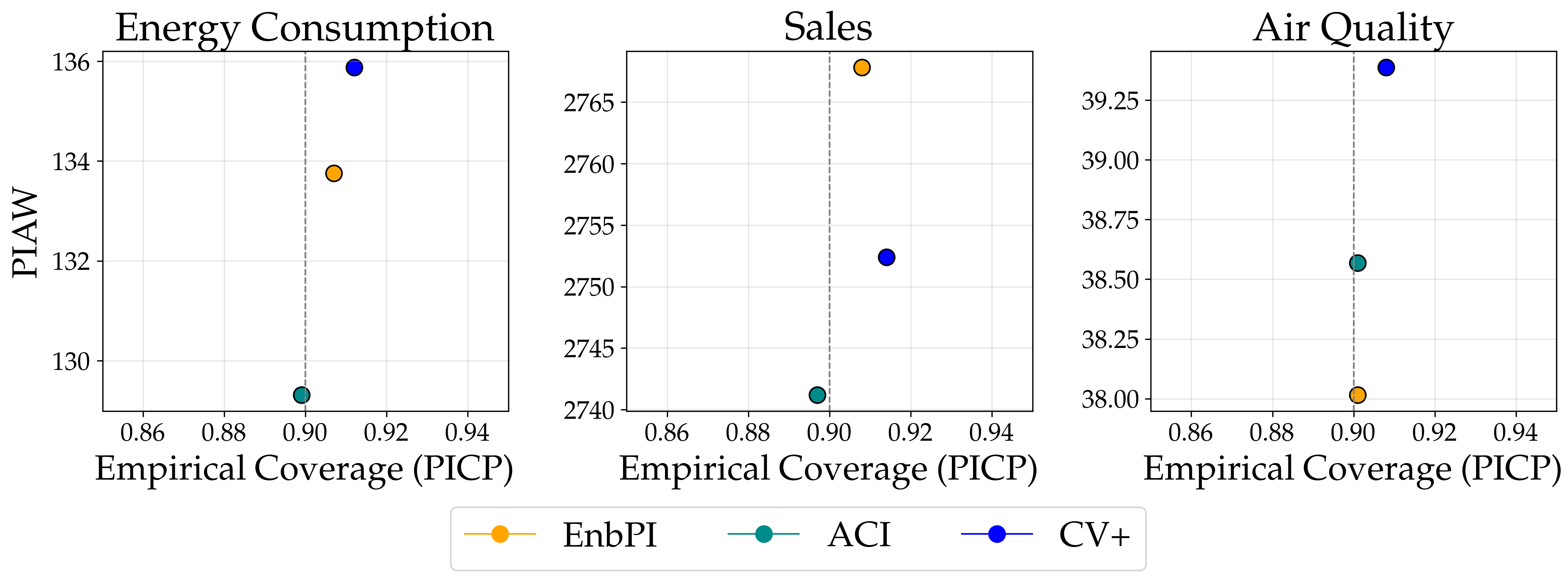}
    \caption{A comparison of decomposition baselines EnbPI and ACI and the CV+ algorithm \citep{barber2020jackknife} across real-world datasets. Metrics are computed based on obtained intervals for the \emph{remainder} component only. The dashed line marks target coverage of $90\%$.}
    \label{fig:remainder-res}
\end{figure}

\paragraph{Remainder component.} As discussed in \autoref{subsec:remainder}, under an effective TSD model the remainder should ideally resemble uncorrelated 'white noise'. Consequently, we investigate whether a method like CV+, which assumes global exchangeability, is practically suitable, or if non-exchangeable approaches that recognize temporal dependencies might be more appropriate. Our results, depicted in \autoref{fig:remainder-res}, confirm that all CP methods reach the intended coverage levels and provide similar prediction interval quality, suggesting that global exchangeability can indeed be sufficient for modelling the remainder term.

\section{Conclusion}
\label{sec:conclusion}

In this work, we investigate the use of conformal prediction methods for time series forecasting through the framework of time series decomposition. We examine both conceptually and empirically the relationship between the components obtained from such a decomposition, and the applicability of different exchangeability regimes that these components may adhere to. By leveraging these exchangeability regimes, we apply specific CP algorithms to each component and combine them to construct prediction intervals that empirically meet conformal guarantees for the overall time series. Our experiments on both synthetic and real-world data indicate that while promising, the effectiveness of obtained uncertainty estimates largely depends on the complexity of the time series, the quality of the decomposition, and other methodological choices. Consequently, our approaches appear most effective for `well-behaved' time series that exhibit high regularity, or when prior knowledge about the data is available to guide modelling decisions. In such settings, closely aligning the employed CP methods with specific exchangeability regimes can be highly beneficial.

Currently, our exploratory investigation focused on an additive decomposition model using the STL algorithm. Future work should address the robustness of our results by \emph{(i)} evaluating more advanced time series decomposition algorithms such as MSTL \citep{bandara2021mstl}, RobustSTL \citep{wen2019robuststl}, or OnlineSTL \citep{mishra2021onlinestl}, which may offer more flexible modelling capabilities; \emph{(ii)} considering different de- and recomposition models, such as for multiplicative structures; and \emph{(iii)} exploring more advanced seasonality weighting schemes, such as adaptive weights that respond to component behaviour, or draw inspiration from local exchangeability conditions seen in \emph{mondrian} CP methods \citep{toccaceli2019combination}. Additionally, open questions remain related to nominal coverage guarantees, and quantifying the impact of decomposition quality on such guarantees. 

By illustrating the potential synergies between conformal prediction and classical time series decomposition, we hope to encourage further research at their intersection, highlighting the critical role of thoughtfully assessing underlying exchangeability assumptions.


\clearpage

\bibliography{references}

\clearpage
\appendix

\section*{Appendix}

The appendix contains additional empirical results and is organized as follows:
\begin{itemize}
    \item \autoref{app:cov-pi}: Comments on the nominal coverage guarantee of decomposition-based approaches.
    \item \autoref{app:res}: Detailed results tables for all considered datasets.
    \item \autoref{app:res-diff-alphas}: Results across varying nominal coverage levels for seasonal CP methods.
    \item \autoref{app:recent-period}: Ablation study on period recency for seasonal CP methods.
    \item \autoref{app:feat-dist}: Ablation study on feature-based weighting for seasonal CP methods.
    \item \autoref{app:decomp}: Qualitative decomposition plots across all datasets.
    \item \autoref{app:rmse}: Results on the predictive performance of employed regression models and hyperparameter settings.
\end{itemize}

\section{Coverage guarantee for the recomposed prediction interval.}
\label{app:cov-pi}

Consider the time series decomposition from \autoref{eq:tsd} and the recomposed prediction interval obtained by applying \autoref{eq:recomp}. Assume that the conformal methods applied to each component (trend $T$, seasonality $S$ and remainder $R$) provide finite-sample, nominal coverage at target level $1 - \alpha$, which we select to be identical for all components ($\alpha_T = \alpha_S = \alpha_R = \alpha$). Then a lower bound on nominal coverage for the recomposed prediction interval $\hat{C}(X_{n+1})$ can be obtained by the union bound as
\begin{equation}
    \begin{split}
        \mathbb{P}(Y_{n+1} &\in \hat{C}(X_{n+1})) \\
        &= \mathbb{P}(T_{n+1} \in \hat{C}_{T}(X_{n+1}) \,\wedge\, S_{n+1} \in \hat{C}_{S}(X_{n+1}) \,\wedge\, R_{n+1} \in \hat{C}_{R}(X_{n+1})) \\
        &= 1 - \mathbb{P}(T_{n+1} \notin \hat{C}_{T}(X_{n+1}) \,\vee\, S_{n+1} \notin \hat{C}_{S}(X_{n+1}) \,\vee\, R_{n+1} \notin \hat{C}_{R}(X_{n+1})) \\
        &\geq 1 - \big[ \underbrace{\mathbb{P}(T_{n+1} \notin \hat{C}_{T}(X_{n+1}))}_{\leq \,\alpha_T} \,+\, \underbrace{\mathbb{P}(S_{n+1} \notin \hat{C}_{S}(X_{n+1}))}_{\leq \,\alpha_S} \,+\, \underbrace{\mathbb{P}(R_{n+1} \notin \hat{C}_{R}(X_{n+1}))}_{\leq \,\alpha_R} \big] \\
        &\geq 1 - 3\,\alpha. \\
    \end{split}
    \label{eq:mht-problem}
\end{equation}
The given bound is loose and does not guarantee target coverage of $1 - \alpha$ for $\hat{C}(X_{n+1})$, an effect that can also be interpreted from a \emph{multiple hypothesis testing} perspective on the conformal procedures applied to each component (see, \eg, \cite{bates2023outlier}). A naive Bonferroni correction can be achieved by selecting $\alpha/3$ as target level per component, but other corrections for the conformal setting have been proposed \citep{sun2022copula, timans2023maxrank}. Since our decomposition approach already produces empirically conservative coverage levels and consists of a combination of different guarantees, we do not evaluate such corrections here. 

\section{Detailed results tables.}
\label{app:res}

We present detailed results tables for our considered methods both on synthetic and real-world experiments. In particular, we report results for a wide range of possible combinations of considered CP methods across decomposition components. That is, we consider combinations of the options $\{$EnbPI, ACI$\}$ for the trend $T$, $\{$EnbPI, ACI, BinaryPoint, BinaryLocal, ExpLocal$\}$ for the season $S$, and $\{$EnbPI, ACI, CV+$\}$ for the remainder $R$. Results are structured by dataset in individual tables. We additionally provide brief dataset descriptions. Overall best results are \underline{underlined}, while best decomposition results are \textbf{boldened}.

\subsection{Synthetic data.}
\label{app:res-synth}

\begin{table}[H]
    \centering
    \renewcommand{\arraystretch}{1}
    \scalebox{0.95}{
    \begin{tabular}{ccccccccc}
    \toprule\toprule
    \multicolumn{3}{c}{\textbf{Conformal algorithm}} & \multicolumn{2}{c}{\textbf{Linear Reg.}} & \multicolumn{2}{c}{\textbf{MLP}} & \multicolumn{2}{c}{\textbf{Gradient Boost.}} \\
    Trend & Season & Remainder & PICP & PIAW & PICP & PIAW & PICP & PIAW \\
    \midrule
    \multicolumn{3}{c}{Original (EnbPI)} & $0.889$ & $41.694$ & $0.887$ & $41.626$ & $0.897$ & $11.922$ \\
    \multicolumn{3}{c}{Original (ACI)}  & $0.898$ & $42.065$ & $0.898$ & $42.168$ & $0.9$ & $11.991$ \\
    \multicolumn{3}{c}{Original (CV+)}  & $0.879$ & $41.645$ & $0.917$ & $42.41$ & $0.904$ & $12.031$ \\
    \midrule
    EnbPI & EnbPI & EnbPI  & $0.985$ & $44.013$ & $0.985$ & $44.197$ & $0.987$ & $12.536$ \\
    EnbPI & BinaryPoint &EnbPI  & $0.944$ & $29.764$ & \underline{\boldsymbol{$0.946$}} & \underline{\boldsymbol{$29.768$}} & \underline{\boldsymbol{$0.948$}} & \underline{\boldsymbol{$9.494$}} \\
    EnbPI & BinaryLocal & EnbPI  & $0.994$ & $41.332$ & $0.994$ & $41.344$ & $0.99$ & $11.913$ \\
    EnbPI & ExpLocal & EnbPI  & $0.994$ & $44.096$ & $0.993$ & $44.005$ & $0.99$ & $12.481$ \\
    EnbPI & EnbPI & CV+ & $0.985$ & $44.003$ & $0.985$ & $44.201$ & $0.988$ & $12.566$ \\
    EnbPI & BinaryPoint & CV+ & \underline{\boldsymbol{$0.946$}} & \underline{\boldsymbol{$29.755$}} & $0.947$ & $29.771$ & $0.947$ & $9.524$ \\
    EnbPI & BinaryLocal & CV+ & $0.994$ & $41.323$ & $0.994$ & $41.347$ & $0.99$ & $11.942$ \\
    EnbPI & ExpLocal & CV+ & $0.994$ & $44.086$ & $0.993$ & $44.009$ & $0.99$ & $12.510$ \\
    \midrule
    ACI & ACI & ACI  & $0.985$ & $43.995$ & $0.986$ & $44.237$ & $0.994$ & $554.298$ \\
    ACI & BinaryPoint & ACI  & $0.946$ & $29.793$ & $0.946$ & $29.822$ & $0.969$ & $551.304$ \\
    ACI & BinaryLocal &ACI  & $0.995$ & $41.361$ & $0.995$ & $41.398$ & $0.996$ & $553.722$ \\
    ACI & ExpLocal & ACI  & $0.994$ & $44.125$ & $0.994$ & $44.06$ & $0.996$ & $554.29$ \\
    ACI & ACI & CV+ & $0.984$ & $43.956$ & $0.985$ & $44.186$ & $0.993$ & $554.321$ \\
    ACI & BinaryPoint & CV+ & \underline{\boldsymbol{$0.946$}} & \underline{\boldsymbol{$29.755$}} & $0.947$ & $29.771$ & $0.968$ & $551.326$ \\
    ACI & BinaryLocal & CV+ & $0.994$ & $41.323$ & $0.994$ & $41.347$ & $0.995$ & $553.744$ \\
    ACI & ExpLocal & CV+ & $0.994$ & $44.086$ & $0.993$ & $44.009$ & $0.995$ & $554.312$ \\
    \bottomrule\bottomrule
    \end{tabular}
    }
    \caption{Detailed results for synthetic data.}
    \label{tab:app-res-synth}
\end{table}

\newpage
\subsection{Real-world data: San Diego energy consumption.}
\label{app:res-energy}

This dataset records energy usage in the city of San Diego, collected over a period of five years. The response consists of hourly energy consumption measurements in Watt hours. Available time-dependent features include seasonal indicators, public holidays, temperature, and air conditioning use, among others.

\begin{table}[H]
    \centering
    \renewcommand{\arraystretch}{1}
    \scalebox{0.95}{
    \begin{tabular}{ccccccccc}
    \toprule\toprule
    \multicolumn{3}{c}{\textbf{Conformal algorithm}} & \multicolumn{2}{c}{\textbf{Linear Reg.}} & \multicolumn{2}{c}{\textbf{MLP}} & \multicolumn{2}{c}{\textbf{Gradient Boost.}} \\
    Trend & Season & Remainder & PICP & PIAW & PICP & PIAW & PICP & PIAW \\
    \midrule
    \multicolumn{3}{c}{Original (EnbPI)} & $0.907$ & $222.866$ & $0.865$ & $161.238$ & $0.899$ & $199.334$ \\
    \multicolumn{3}{c}{Original (ACI)} &  \underline{$0.9$} & \underline{$218.301$} & $0.9$ & $187.367$ & $0.9$ & $204.007$ \\
    \multicolumn{3}{c}{Original (CV+)} & $0.912$ & $227.846$ & $0.898$ & $168.878$ & \underline{$0.9$} & \underline{$198.872$} \\
    \midrule
    EnbPI &EnbPI &EnbPI & $0.964$ & $243.005$ & \underline{\boldsymbol{$0.944$}} & \underline{\boldsymbol{$161.773$}} & $0.966$ & $271.898$ \\
    EnbPI &BinaryPoint &EnbPI & $0.97$ & $232.852$ & $0.961$ & $172.657$ & $0.973$ & $272.751$ \\
    EnbPI &BinaryLocal& EnbPI & $0.968$ & $238.593$ & $0.963$ & $177.651$ & $0.977$ & $281.453$ \\
    EnbPI &ExpLocal& EnbPI & $0.968$ & $238.113$ & $0.963$ & $176.555$ & $0.977$ & $281.194$ \\
    EnbPI &EnbPI &CV+ & $0.963$ & $245.122$ & $0.956$ & $170.989$ & $0.967$ & $273.554$ \\
    EnbPI &BinaryPoint &CV+ & $0.971$ & $234.969$ & $0.972$ & $181.872$ & $0.975$ & $274.407$ \\
    EnbPI &BinaryLocal &CV+ & $0.967$ & $240.71$ & $0.974$ & $186.867$ & $0.977$ & $283.108$ \\
    EnbPI &ExpLocal &CV+ & $0.967$ & $240.23$ & $0.974$ & $185.77$ & $0.977$ & $282.849$ \\
    \midrule
    ACI &ACI &ACI & $0.96$ & $229.425$ & $0.966$ & $190.995$ & $0.968$ & $278.927$ \\
    ACI &BinaryPoint &ACI & \boldsymbol{$0.969$} & \boldsymbol{$228.552$} & $0.969$ & $181.361$ & \boldsymbol{$0.97$} & \boldsymbol{$270.637$} \\
    ACI &BinaryLocal &ACI & $0.967$ & $234.293$ & $0.97$ & $186.355$ & $0.974$ & $279.338$ \\
    ACI &ExpLocal &ACI & $0.968$ & $233.814$ & $0.97$ & $185.259$ & $0.973$ & $279.07$ \\
    ACI &ACI &CV+ & $0.962$ & $235.855$ & $0.971$ & $191.174$ & $0.972$ & $286.686$ \\
    ACI &BinaryPoint &CV+ & $0.971$ & $234.983$ & $0.971$ & $181.54$ & $0.974$ & $278.396$ \\
    ACI &BinaryLocal &CV+ & $0.967$ & $240.724$ & $0.973$ & $186.534$ & $0.977$ & $287.097$ \\
    ACI &ExpLocal &CV+ & $0.967$ & $240.245$ & $0.973$ & $185.438$ & $0.977$ & $286.838$ \\
    \bottomrule\bottomrule
    \end{tabular}
    }
    \caption{Detailed results for San Diego energy consumption data.}
    \label{tab:app-res-energy}
\end{table}

\newpage
\subsection{Real-world data: Rossman store sales.}
\label{app:res-sales}

This dataset contains aggregated daily sales from the chain of Rossmann drug stores. The response consists of average daily sales numbers. Available time-dependent features include weekday, average number of customers, and whether holiday indicators.

\begin{table}[H]
    \centering
    \renewcommand{\arraystretch}{1}
    \scalebox{0.95}{
    \begin{tabular}{ccccccccc}
    \toprule\toprule
    \multicolumn{3}{c}{\textbf{Conformal algorithm}} & \multicolumn{2}{c}{\textbf{Linear Reg.}} & \multicolumn{2}{c}{\textbf{MLP}} & \multicolumn{2}{c}{\textbf{Gradient Boost.}} \\
    Trend & Season & Remainder & PICP & PIAW & PICP & PIAW & PICP & PIAW \\
    \midrule
    \multicolumn{3}{c}{Original (EnbPI)}   & $0.892$ & $1572.343$ & $0.923$ & $1231.569$ & $0.932$ & $5476.084$ \\
    \multicolumn{3}{c}{Original (ACI)} & \underline{$0.895$} & \underline{$1539.925$} & \underline{$0.906$} & \underline{$1172.582$} & \underline{$0.919$} & \underline{$4867.123$} \\
    \multicolumn{3}{c}{Original (CV+)} & $0.892$ & $1512.231$ & $0.928$ & $1276.386$ & $0.924$ & $5470.683$ \\
    \midrule
    EnbPI & EnbPI & EnbPI & $1.0$ & $5303.825$ & $0.975$ & $4870.155$ & \boldsymbol{$0.932$} & \boldsymbol{$6075.189$} \\
    EnbPI & BinaryPoint & EnbPI & $1.0$ & $6060.792$ & $0.989$ & $6404.697$ & $0.978$ & $8051.558$ \\
    EnbPI & BinaryLocal & EnbPI & $1.0$ & $5441.858$ & $0.984$ & $5553.258$ & $0.973$ & $7547.193$ \\
    EnbPI & ExpLocal & EnbPI & $1.0$ & $5263.612$ & $0.984$ & $5381.6$ & $0.658$ & $7373.728$ \\
    EnbPI & EnbPI & CV+ & $1.0$ & $5288.406$ & $0.986$ & $5297.233$ & $0.935$ & $6284.836$ \\
    EnbPI & BinaryPoint & CV+ & $1.0$ & $6045.373$ & $1.0$ & $6831.775$ & $0.98$ & $8261.205$ \\
    EnbPI & BinaryLocal & CV+ & $1.0$ & $5426.439$ & $0.996$ & $5980.336$ & $0.968$ & $7756.839$ \\
    EnbPI & ExpLocal & CV+ & $1.0$ & $5248.193$ & $0.996$ & $5808.679$ & $0.976$ & $7583.375$ \\
    \midrule
    ACI & ACI & ACI & $1.0$ & $5447.39$ & \boldsymbol{$0.984$} & \boldsymbol{$4615.531$} & $0.944$ & $6373.45$ \\
    ACI & BinaryPoint & ACI & $1.0$ & $6015.128$ & $0.993$ & $6313.72$ & $0.977$ & $8003.03$ \\
    ACI & BinaryLocal & ACI & $1.0$ & $5396.194$ & $0.987$ & $5462.281$ & $0.973$ & $7498.665$ \\
    ACI & ExpLocal & ACI & \boldsymbol{$1.0$} & \boldsymbol{$5217.948$} & $0.974$ & $5290.623$ & $0.966$ & $7325.201$ \\
    ACI & ACI & CV+ & $1.0$ & $5477.696$ & $0.989$ & $5201.327$ & $0.944$ & $6623.43$ \\
    ACI & BinaryPoint & CV+ & $1.0$ & $6045.434$ & $1.0$ & $6899.516$ & $0.98$ & $8253.01$ \\
    ACI & BinaryLocal & CV+ & $1.0$ & $5426.5$ & $0.996$ & $6048.076$ & $0.968$ & $7748.645$ \\
    ACI & ExpLocal & CV+ & $1.0$ & $5248.254$ & $0.996$ & $5876.419$ & $0.968$ & $7575.181$ \\
    \bottomrule\bottomrule
    \end{tabular}
    }
    \caption{Detailed results for Rossman store sales data.}
    \label{tab:app-res-sales}
\end{table}

\newpage 
\subsection{Real-world data: Beijing air quality.}
\label{app:res-air}

This dataset consists of hourly readings of air pollutants and meteorological data from the city of Beijing. The primary response is the concentration of PM2.5 pollutants in the air. Available time-dependent features include various pollutant concentrations (\eg, SO2, NO2, CO, O3); air temperature and pressure; or rainfall, wind direction and speed. 

\begin{table}[H]
    \centering
    \renewcommand{\arraystretch}{1}
    \scalebox{0.95}{
    \begin{tabular}{ccccccccc}
    \toprule\toprule
    \multicolumn{3}{c}{\textbf{Conformal algorithm}} & \multicolumn{2}{c}{\textbf{Linear Reg.}} & \multicolumn{2}{c}{\textbf{MLP}} & \multicolumn{2}{c}{\textbf{Gradient Boost.}} \\
    Trend & Season & Remainder & PICP & PIAW & PICP & PIAW & PICP & PIAW \\
    \midrule
    \multicolumn{3}{c}{Original (EnbPI)} & \underline{$0.902$} & \underline{$42.918$} & $0.892$ & $40.098$ & \underline{$0.905$} & \underline{$44.889$} \\
    \multicolumn{3}{c}{Original (ACI)} & $0.901$ & $43.549$ & $0.901$ & $44.733$ & $0.901$ & $46.435$ \\
    \multicolumn{3}{c}{Original (CV+)} &  $0.908$ & $44.499$ & \underline{$0.915$} & \underline{$44.398$} & $0.913$ & $46.909$ \\
    \midrule
    EnbPI & EnbPI & EnbPI & $0.954$ & $62.721$ & $0.955$ & $64.756$ & $0.951$ & $64.74$ \\
    EnbPI & BinaryPoint & EnbPI & $0.956$ & $65.643$ & $0.959$ & $68.696$ & $0.952$ & $68.127$ \\
    EnbPI & BinaryLocal & EnbPI & $0.953$ & $61.345$ & $0.956$ & $64.376$ & $0.948$ & $63.489$ \\
    EnbPI & ExpLocal & EnbPI & \boldsymbol{$0.953$} & \boldsymbol{$60.869$} & \boldsymbol{$0.956$} & \boldsymbol{$63.69$} & \boldsymbol{$0.964$} & \boldsymbol{$62.882$} \\
    EnbPI & EnbPI & CV+ & $0.956$ & $64.092$ & $0.962$ & $67.586$ & $0.954$ & $66.537$ \\
    EnbPI & BinaryPoint & CV+ & $0.958$ & $67.014$ & $0.963$ & $71.526$ & $0.954$ & $69.924$ \\
    EnbPI & BinaryLocal & CV+ & $0.955$ & $62.716$ & $0.96$ & $67.206$ & $0.951$ & $65.286$ \\
    EnbPI & ExpLocal & CV+ & $0.954$ & $62.24$ & $0.96$ & $66.52$ & $0.950$ & $64.68$ \\
    \midrule
    ACI & ACI & ACI & $0.958$ & $63.445$ & $0.961$ & $67.267$ & $0.957$ & $74.507$ \\
    ACI & BinaryPoint & ACI & $0.962$ & $66.195$ & $0.965$ & $70.108$ & $0.96$ & $76.922$ \\
    ACI & BinaryLocal & ACI & $0.957$ & $61.897$ & $0.961$ & $65.789$ & $0.958$ & $72.284$ \\
    ACI & ExpLocal & ACI & $0.957$ & $61.421$ & $0.961$ & $65.103$ & $0.976$ & $71.678$ \\
    ACI & ACI & CV+ & $0.958$ & $64.264$ & $0.965$ & $68.233$ & $0.961$ & $74.603$ \\
    ACI & BinaryPoint & CV+ & $0.958$ & $67.014$ & $0.963$ & $71.074$ & $0.956$ & $77.019$ \\
    ACI & BinaryLocal & CV+ & $0.955$ & $62.716$ & $0.96$ & $66.755$ & $0.953$ & $72.381$ \\
    ACI & ExpLocal & CV+ & $0.954$ & $62.24$ & $0.959$ & $66.068$ & $0.953$ & $71.774$ \\
        \bottomrule\bottomrule
        \end{tabular}
    }
    \caption{Detailed results for Beijing air quality data.}
    \label{tab:app-res-air}
\end{table}

\newpage
\section{Results across varying coverage levels.}
\label{app:res-diff-alphas}

\begin{figure}[H]
    \centering
    \includegraphics[width=0.8\textwidth]{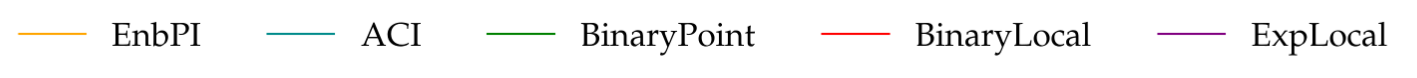}
    \\
    \includegraphics[width=0.8\textwidth]{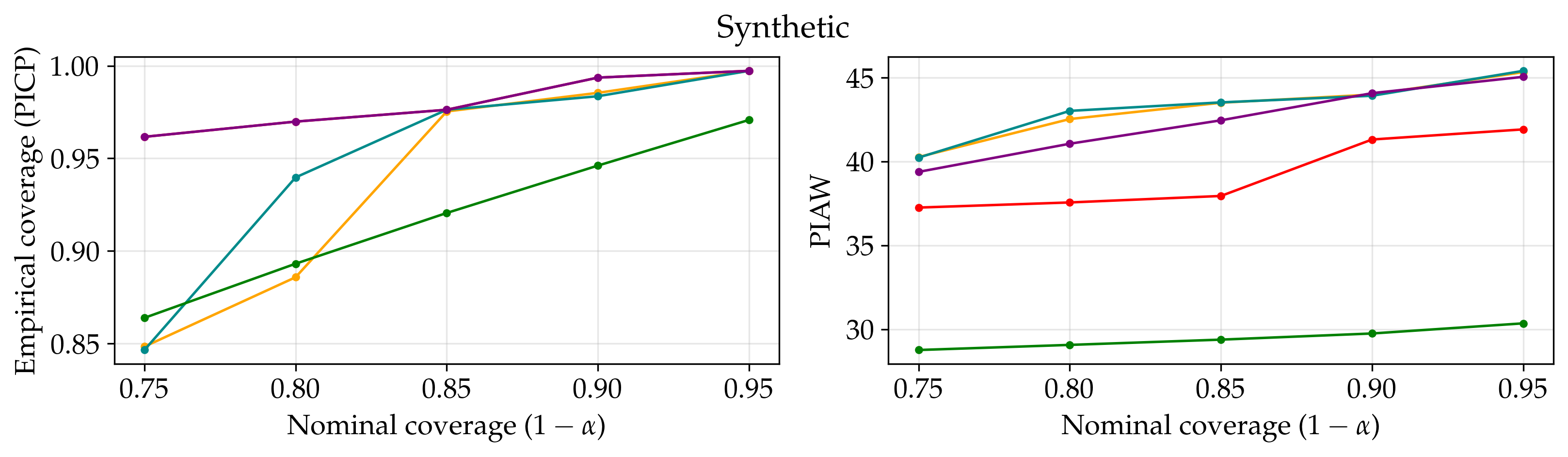}
    \\
    \includegraphics[width=0.8\textwidth]{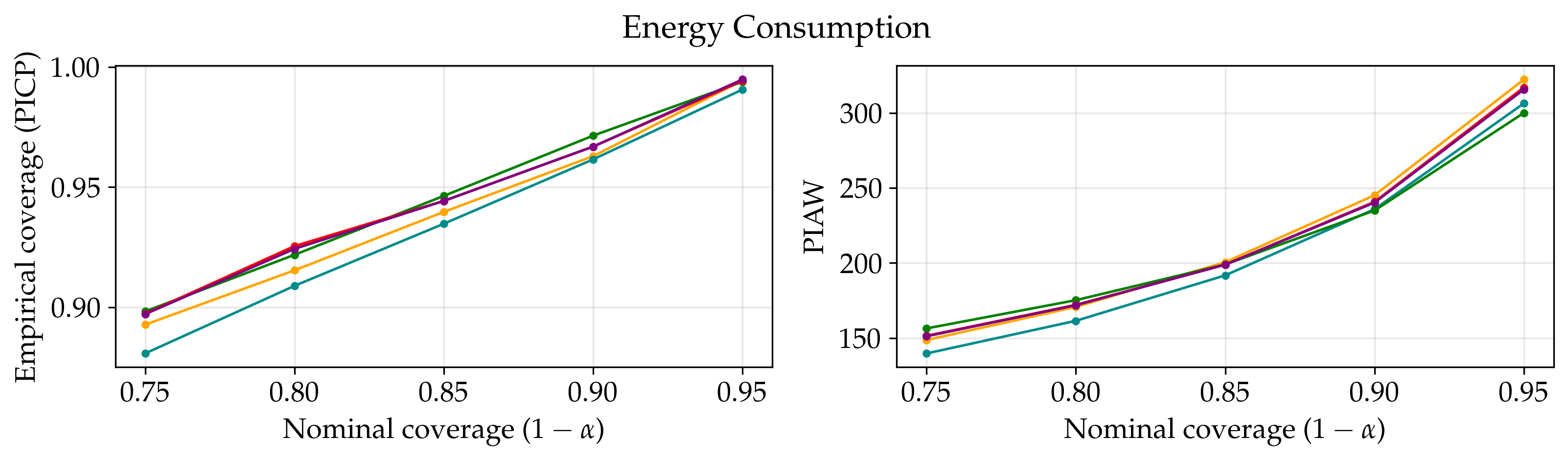}
    \\
    \includegraphics[width=0.8\textwidth]{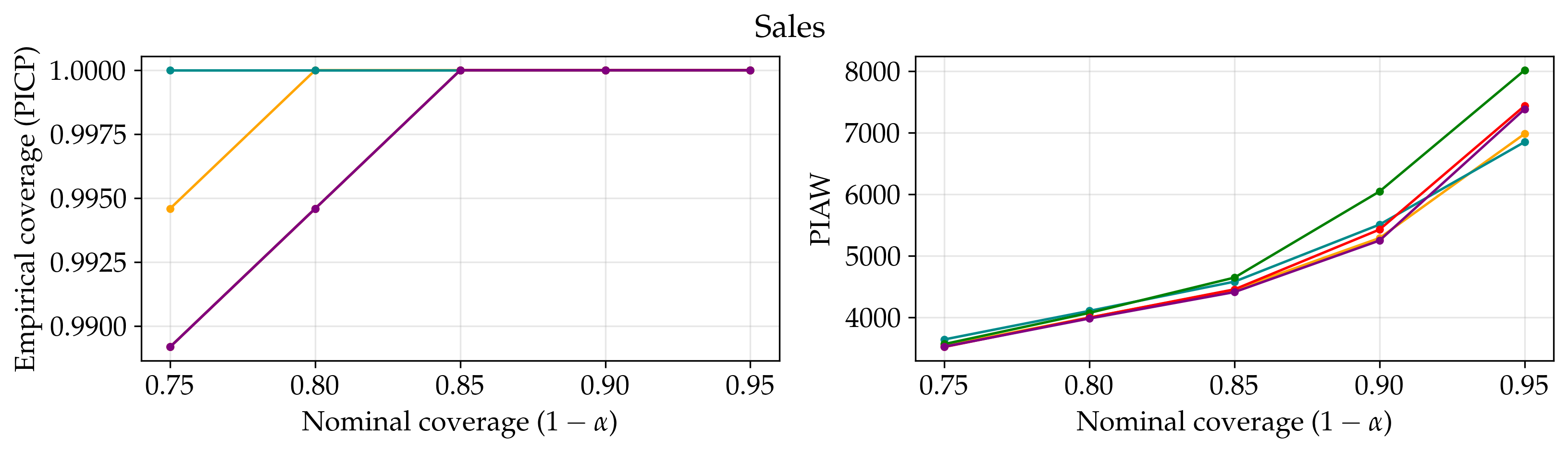}
    \\
    \includegraphics[width=0.8\textwidth]{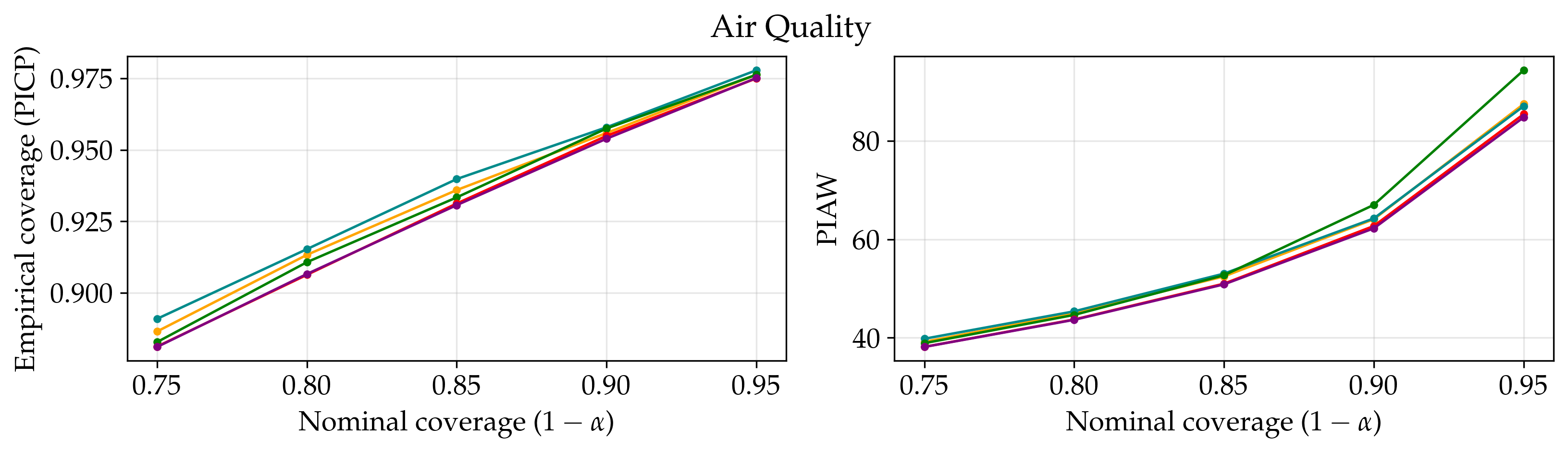}
    \caption{Results for varying seasonal CP methods across nominal coverage levels.} 
    \label{fig:app-res-alphas}
\end{figure}

We validate our results from \autoref{tab:synth_samp} and \autoref{tab:real_samp} across varying nominal coverage levels $(1-\alpha) \in \{0.75, 0.8, 0.85, 0.9, 0.95\}$ in \autoref{fig:app-res-alphas}. Once again we consider the linear regressor with CP methods EnbPI and CV+ for trend and remainder terms, and vary our approaches for the seasonal term. We continue to observe that overcoverage persists across all coverage levels and datasets. For synthetic data, our weighting schemes (in particular BinaryPoint) consistently outperform other methods. For real-world data, methods show similar performance across coverage levels, with relatively small deviations. The general trends in terms of empirical coverage and interval width across nominal coverage levels are in line with expectations, \ie, higher coverage requirements result in both larger PICP and PIAW. 

\newpage
\section{Varying the number of selected periods based on recency.}
\label{app:recent-period}

\begin{figure}[H]
    \centering
    \includegraphics[width=0.75\textwidth]{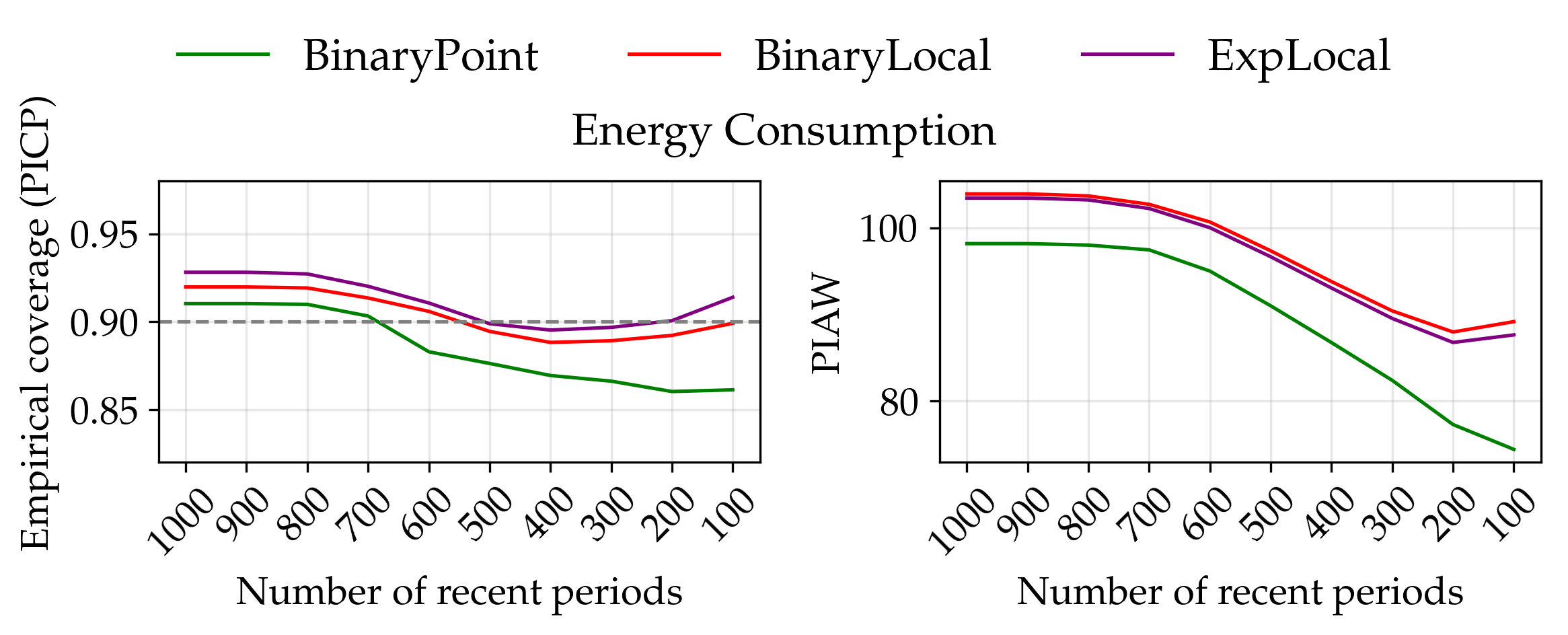}
    \\
    \includegraphics[width=0.75\textwidth]{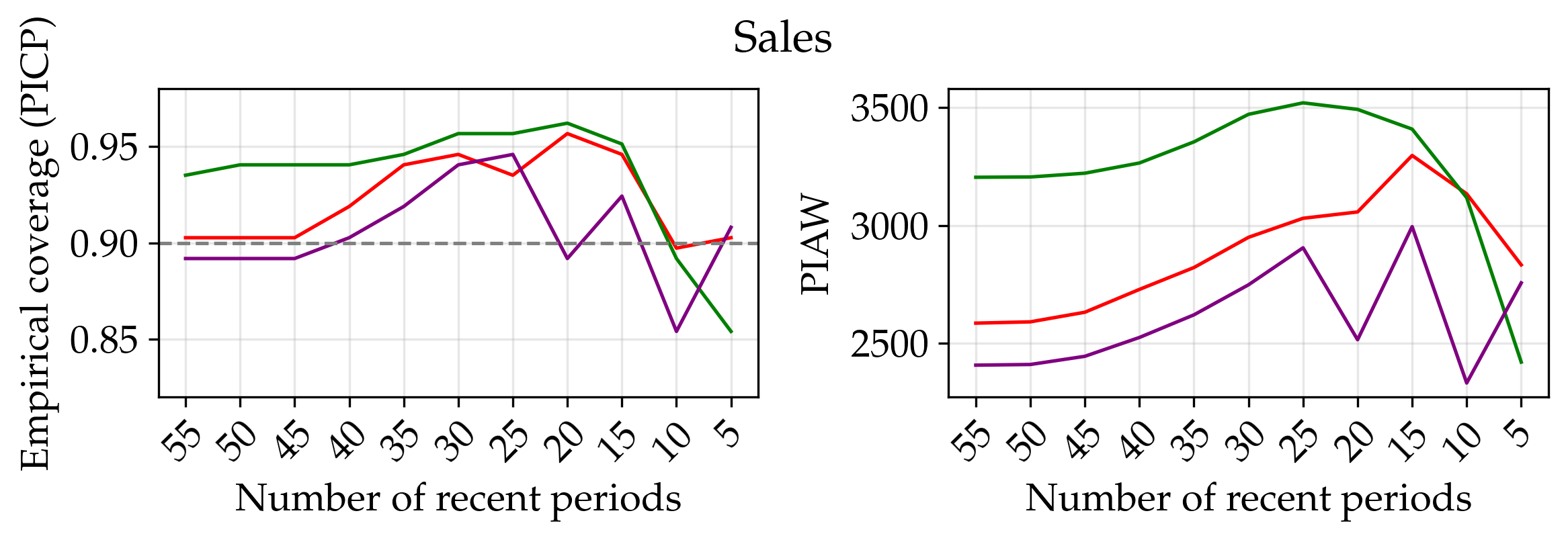}
    \\
    \includegraphics[width=0.75\textwidth]{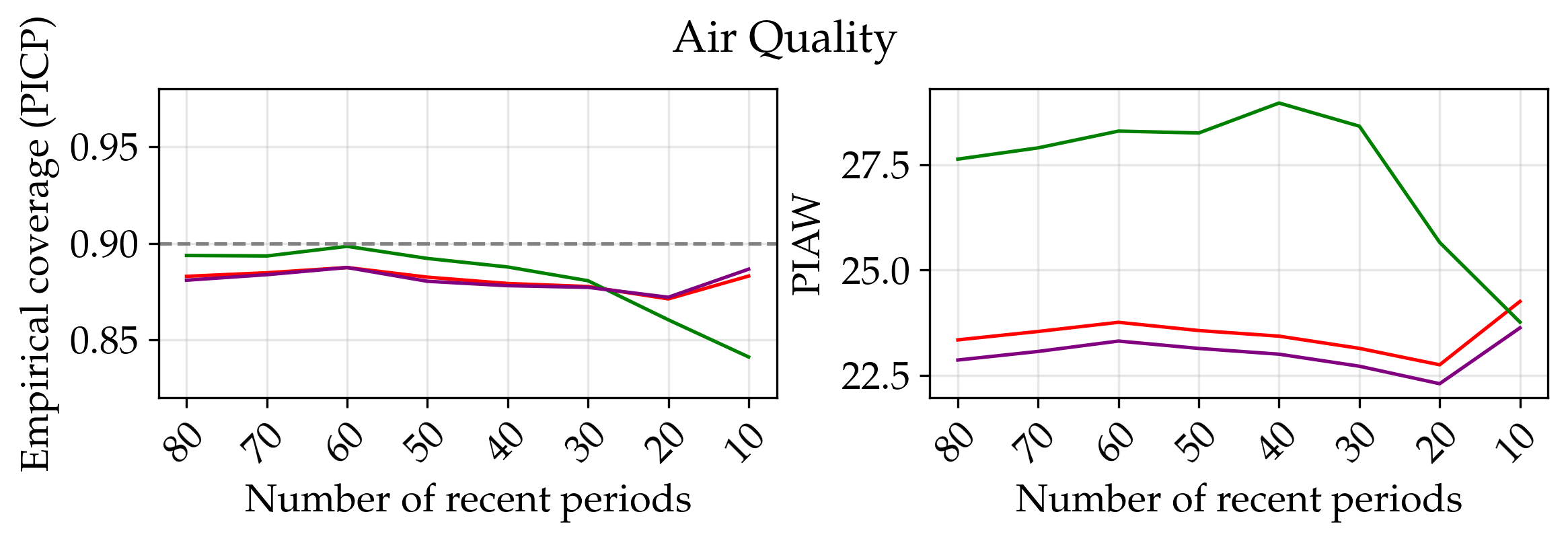}
    \caption{Ablation on temporal proximity of periods on real-world data.}
    \label{fig:app-recency}
\end{figure}

We conduct an ablation study to assess how results are affected by calibration set sizes in terms of temporal proximity in \autoref{fig:app-recency}. That is, for a given seasonal weighting mechanism (BinaryPoint, BinaryLocal or ExpLocal) we consider only leveraging a subset of the most recent periods in time, inspired by the recency-based weighting schemes employed in \cite{barber2023conformal}. We observe that while BinaryPoint experiences generally narrow PIs when fewer periods are used, its empirical coverage also decreases below the target level of $90\%$. In contrast, both BinaryLocal and ExpLocal maintain coverage closer to the desired level throughout, whilst occasionally displaying more efficient PIs. Overall, the results across datasets are mixed, but indicate that temporal proximity can indeed play a role in further reducing interval sizes whilst guaranteeing target coverage. Since slow-moving trend or seasonality shifts might occur over long periods of time, discarding time-distant observations can be sensible.

\section{Seasonal weighting mechanisms based on feature distance.}
\label{app:feat-dist}

Related to the approach by \cite{guan2023localized}, we also consider weighting schemes for the seasonal component which are data-dependent and based on feature distances. We consider two approaches which weigh calibration samples proportional to the euclidean feature distance from the test sample, one with clustering (\textbf{KNN}) and one without (\textbf{FeatDistPoint}). Our ablation is performed on the air quality dataset, which is well-suited for feature-based methods due to its large set of natural features (beyond lags).

\begin{figure}[H]
    \centering
    \begin{minipage}[t]{0.45\textwidth}
        \centering
        \includegraphics[width=\textwidth]{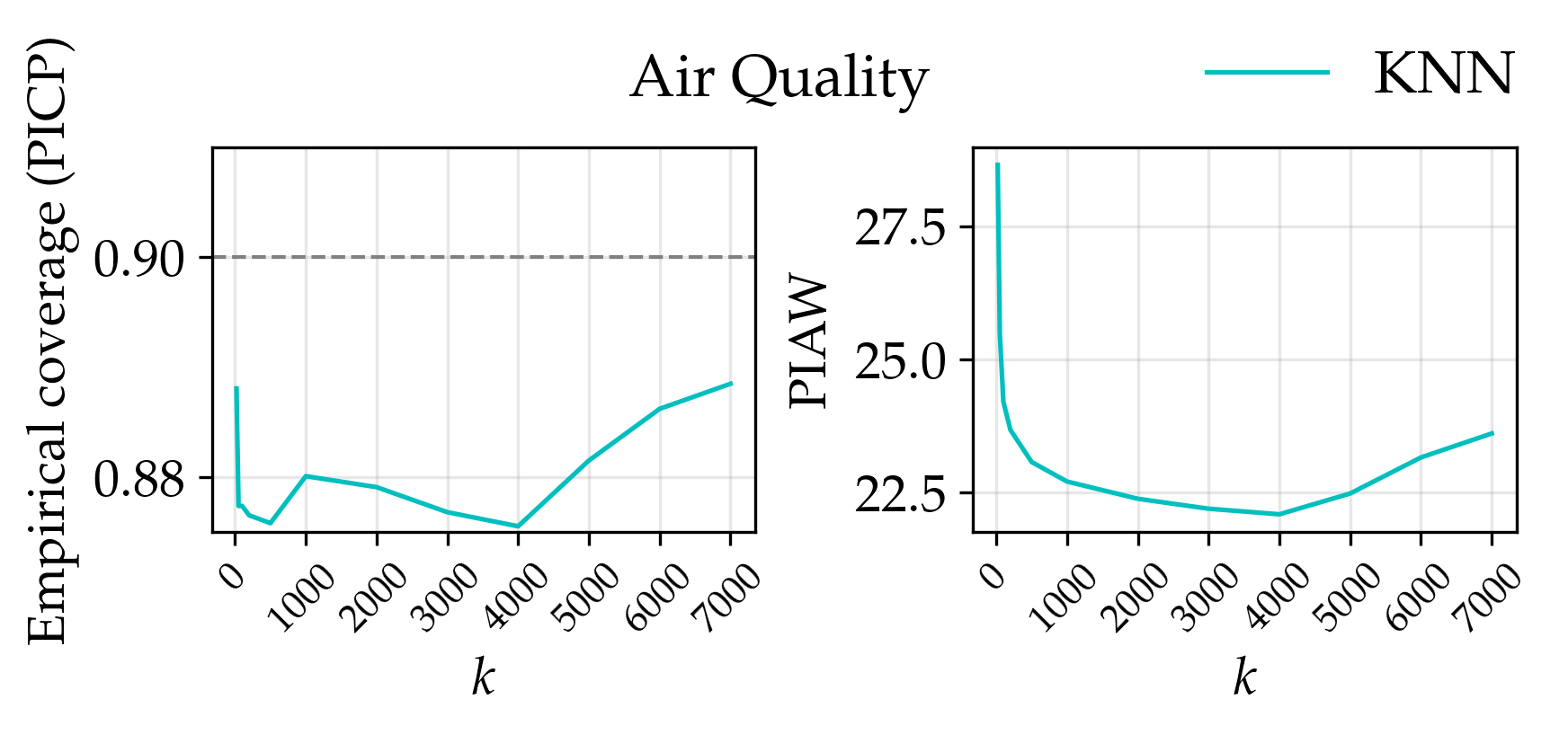}
        \caption{The effect of varying the neighbourhood size $k$ for KNN.}
        \label{fig:knn_ablation}
    \end{minipage}
    \hfill
    \begin{minipage}[t]{0.5\textwidth}
        \centering
        \includegraphics[width=\textwidth]{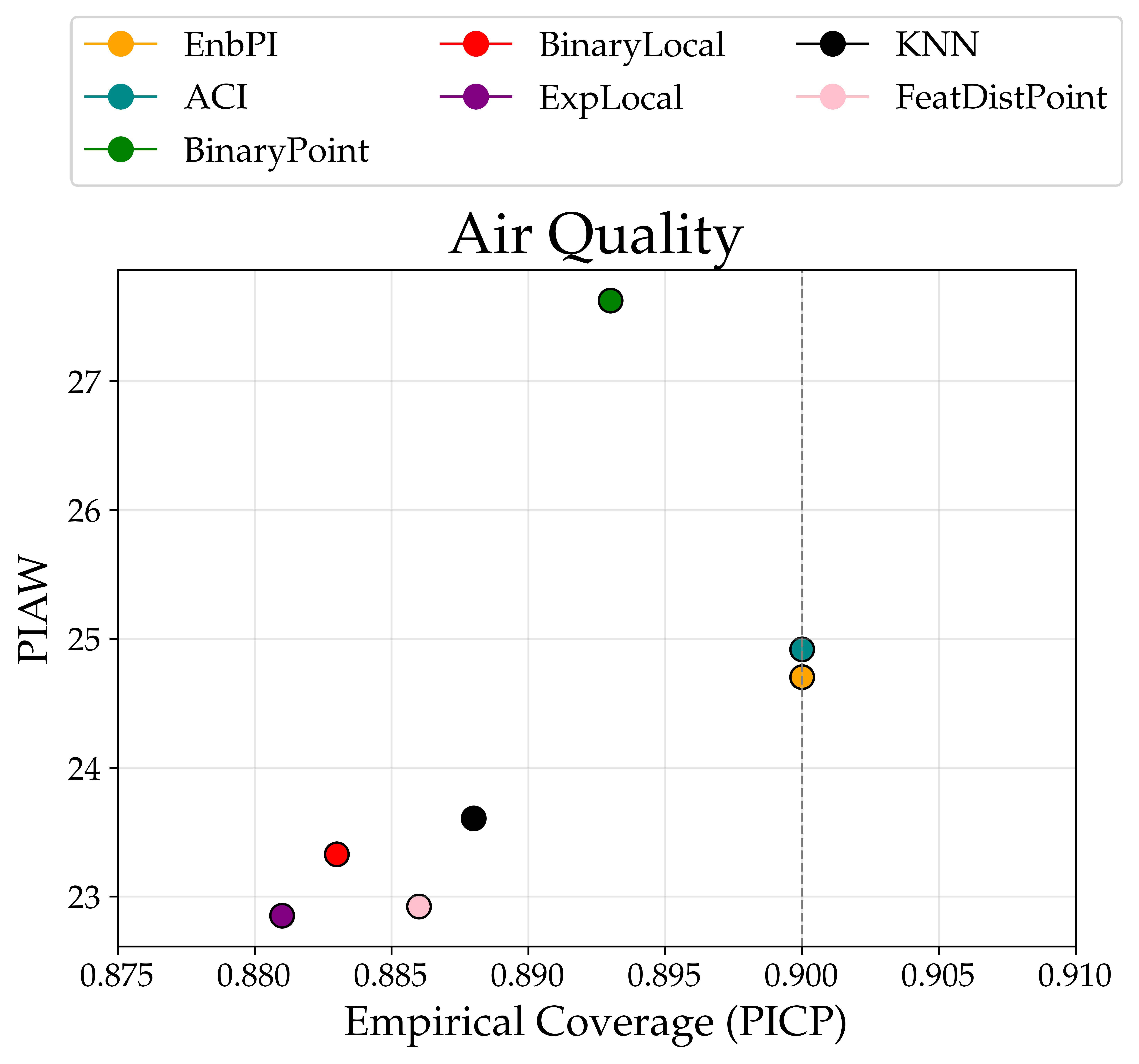}
        \caption{Comparison of CP methods for the \emph{seasonal component} on Air Quality data.}
        \label{fig:air_quality_seasonal_feat_dist}
    \end{minipage}
\end{figure}

\paragraph{KNN.} For the first approach we employ the $k$-nearest neighbours algorithm (KNN), computing the feature distances $d_i := \Vert X_{n+1} - X_i \Vert_2\, \,\,\forall i \in \D_{cal}$ and assigning the $k$-th nearest neighbours an unnormalized weight of one, while any other calibration samples are discarded. This provides a feature-based binary weighting scheme, similar to BinaryPoint. We investigate the effect of choosing the neighbourhood size $k$ in \autoref{fig:knn_ablation}, observing a mixed impact on coverage but a reduction in interval size when only about $50\%$ of available samples ($k=4000$) are selected. 

\paragraph{FeatDistPoint.} Our second approach provides a softer weighting scheme similar to ExpLocal, wherein each calibration sample is assigned a weight inversely proportional to its distance (\ie, $\tilde{w}_i = 1/d_i$), ensuring that samples closer to $X_{n+1}$ have larger weights in \autoref{eq:weight-cp}. \\

\noindent We compare our feature-based weighting schemes against EnbPI, ACI, and approaches based on temporal distance (\autoref{subsec:season}) in \autoref{fig:air_quality_seasonal_feat_dist} (we select $k=7000$ for KNN for maximal coverage). Feature-based weights perform similarly well to other approaches, suggesting that leveraging the feature domain is a viable option. However, this may come at the cost of potentially more involved computations in its generally higher dimensionality.

\section{Qualitative decomposition plots across datasets.}
\label{app:decomp}

\begin{figure}[H]
    \includegraphics[width=0.32\textwidth]{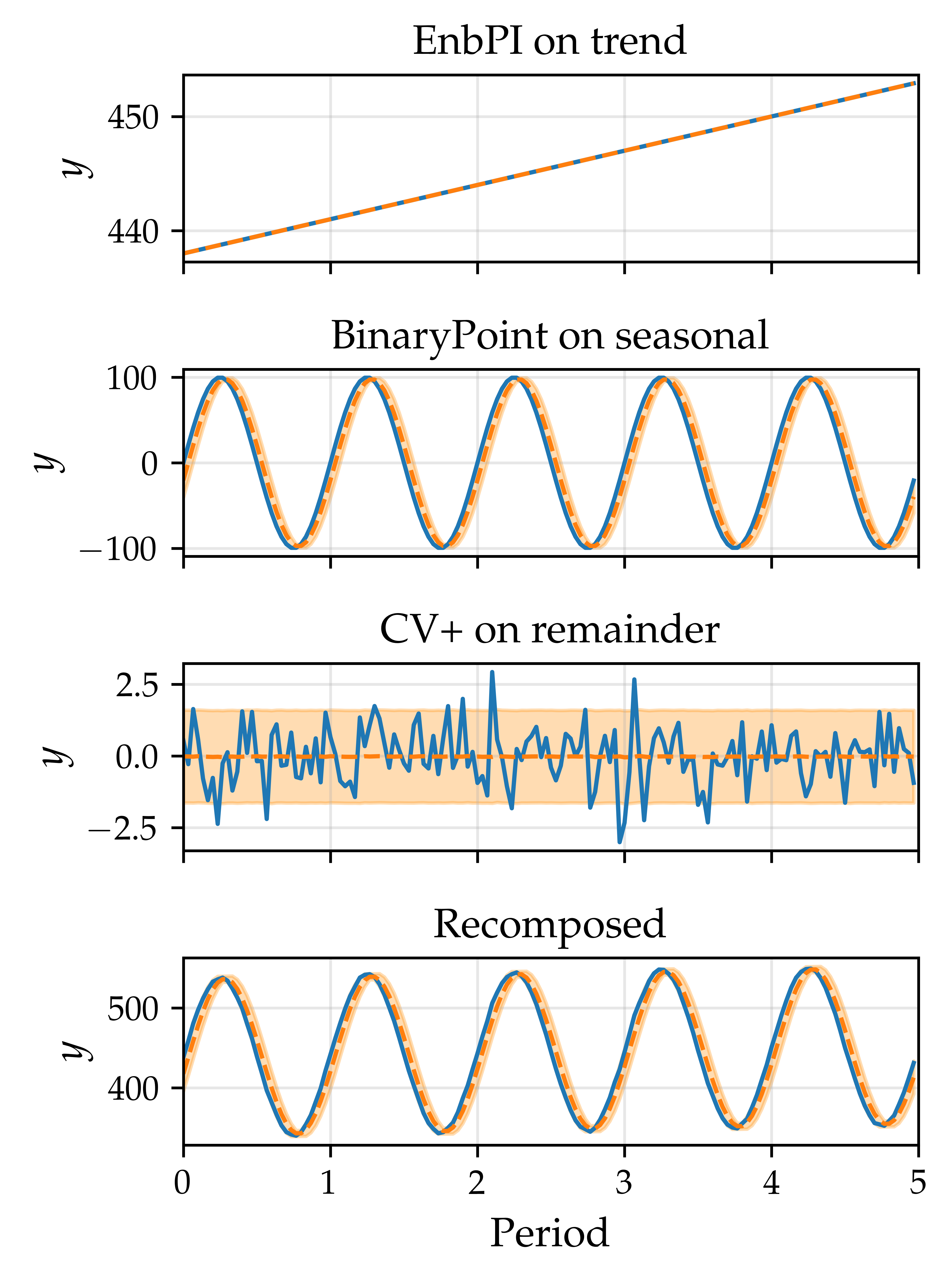}
    \includegraphics[width=0.32\textwidth]{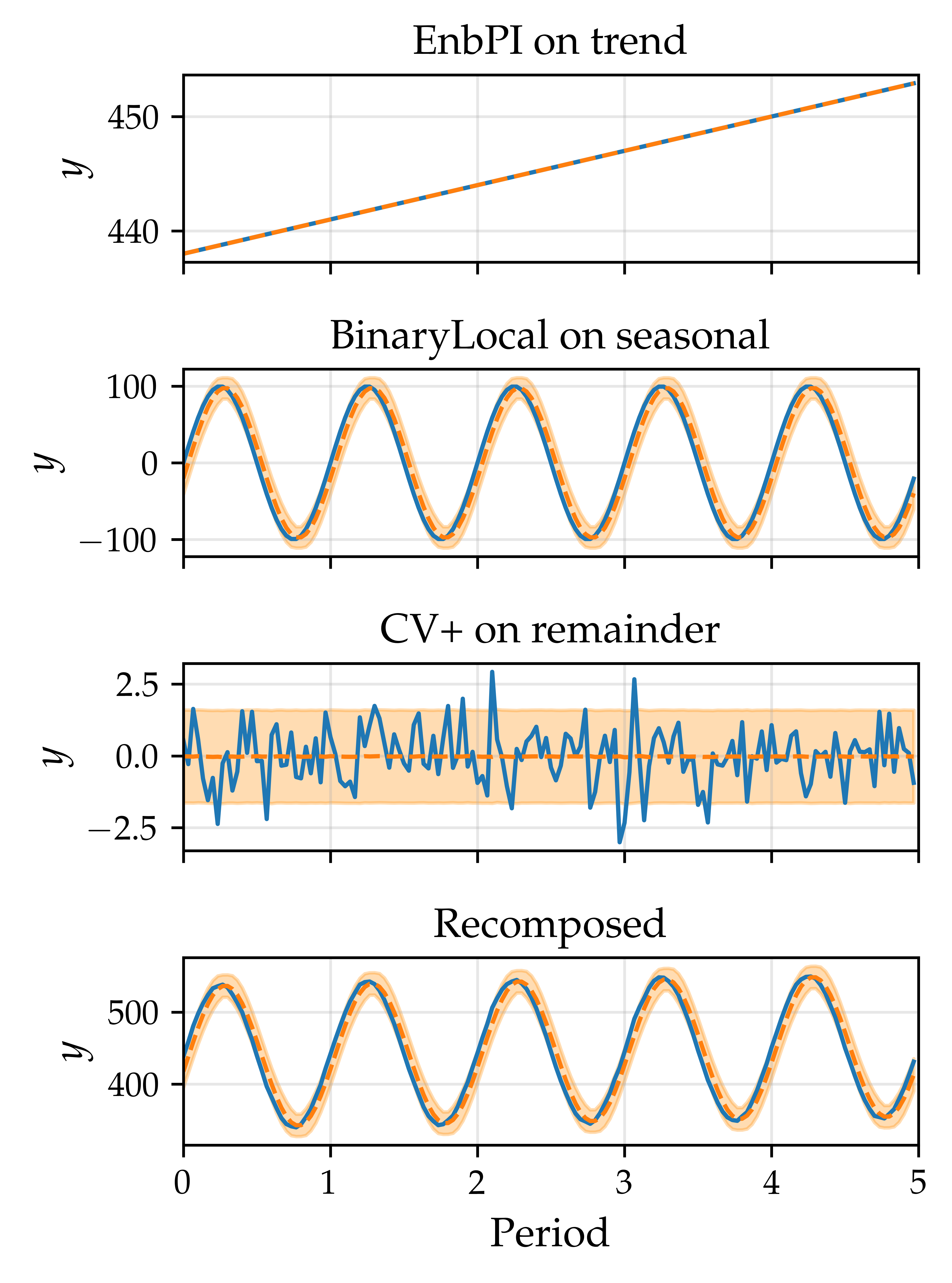}
    \includegraphics[width=0.32\textwidth]{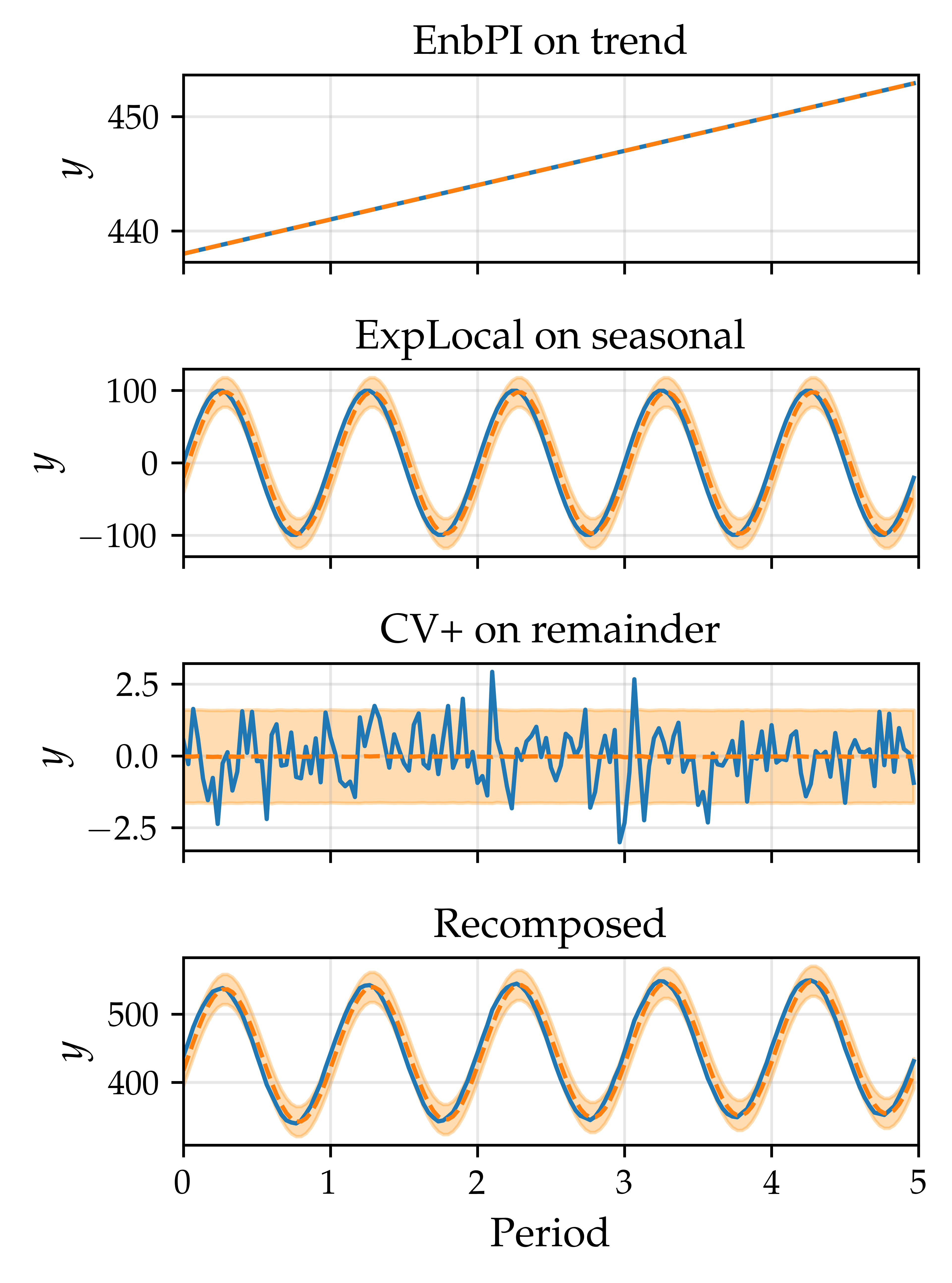}
    \caption{Time series and conformal prediction intervals per component and for the recomposition on \textbf{synthetic} data. \emph{From left to right}: The seasonal component leverages BinaryPoint, BinaryLocal or ExpLocal.}
    \label{fig:decomposition_synthetic}
\end{figure}

\begin{figure}[H]
    \includegraphics[width=0.32\textwidth]{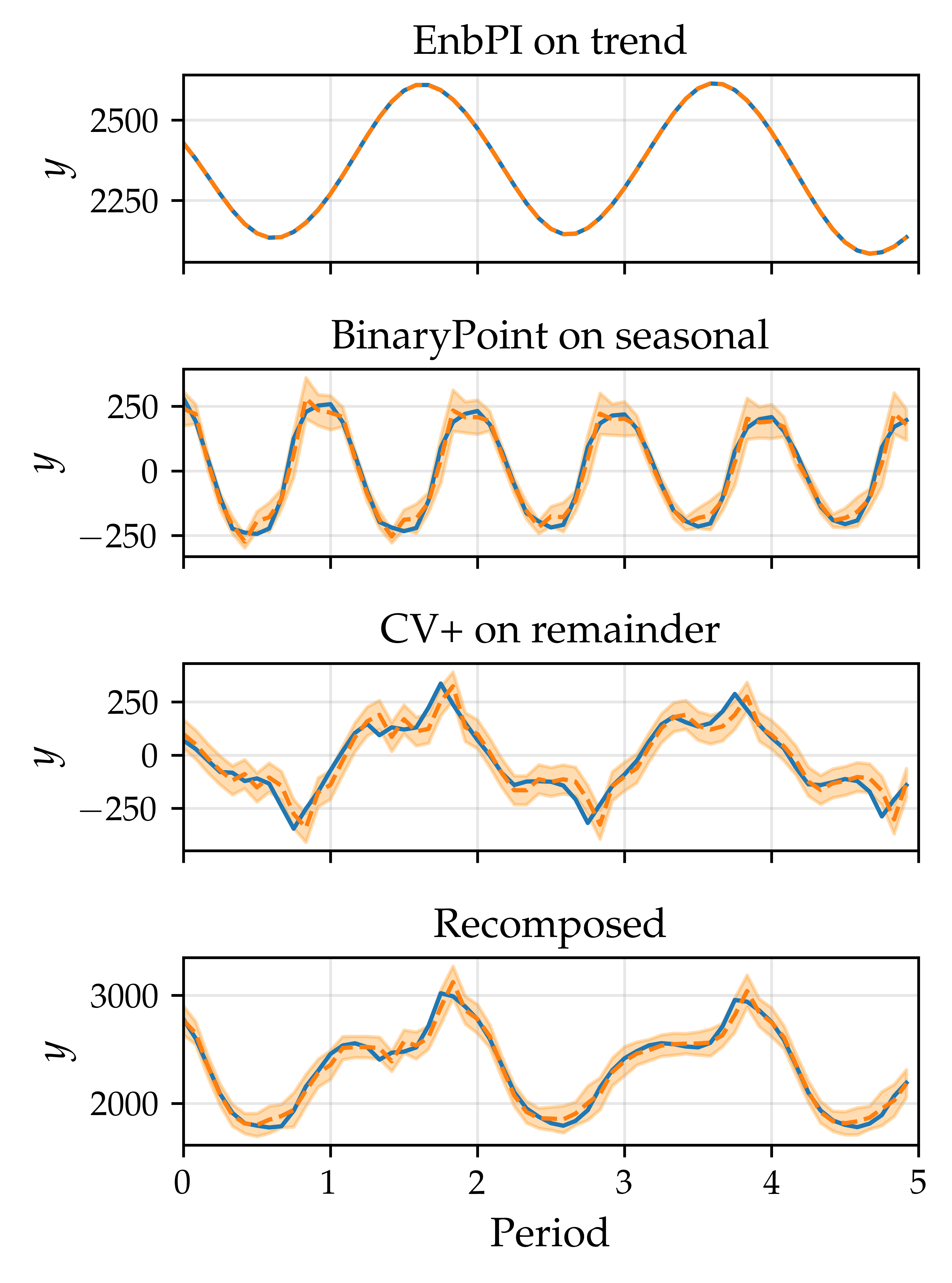}
    \includegraphics[width=0.32\textwidth]{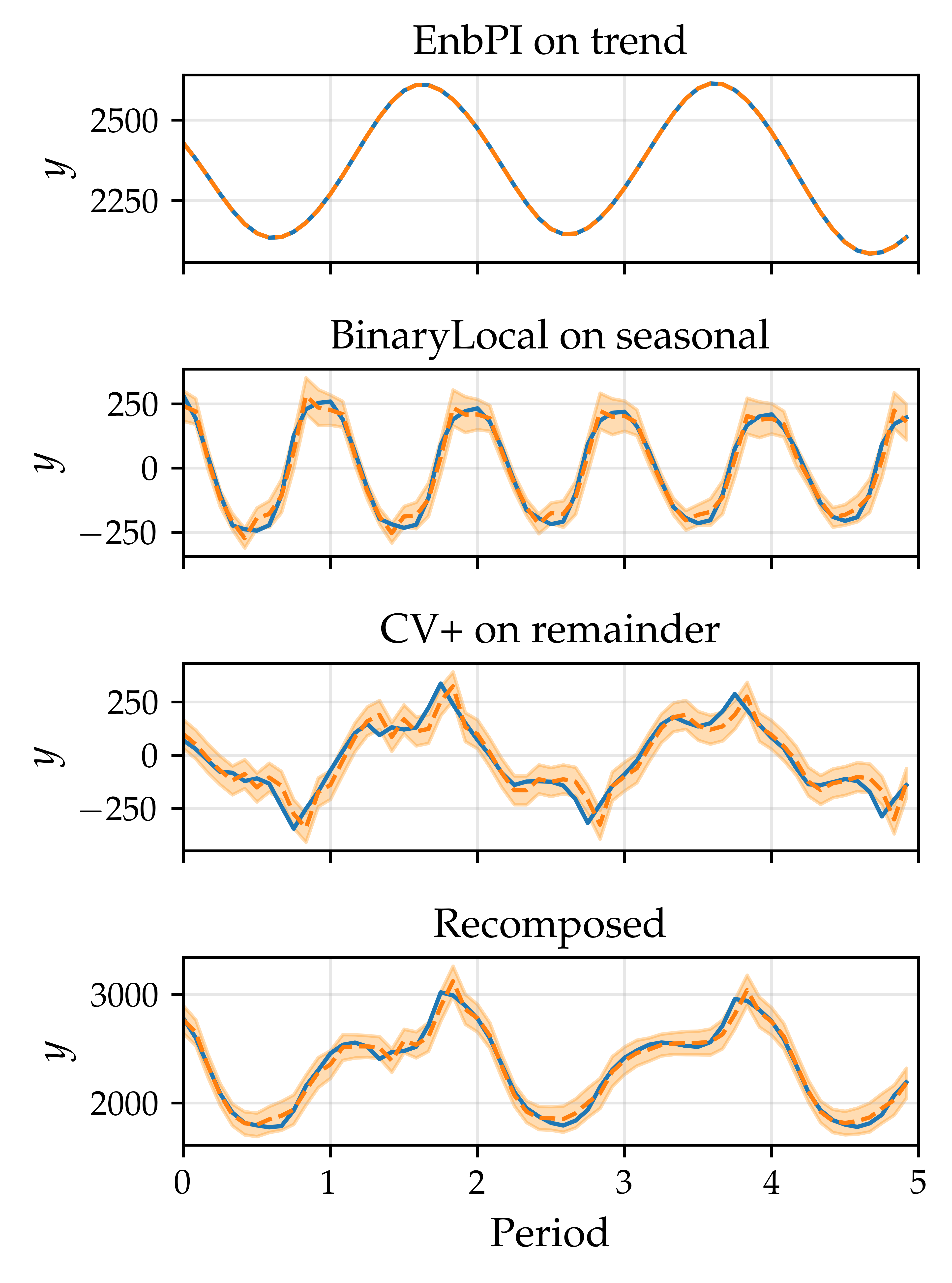}
    \includegraphics[width=0.32\textwidth]{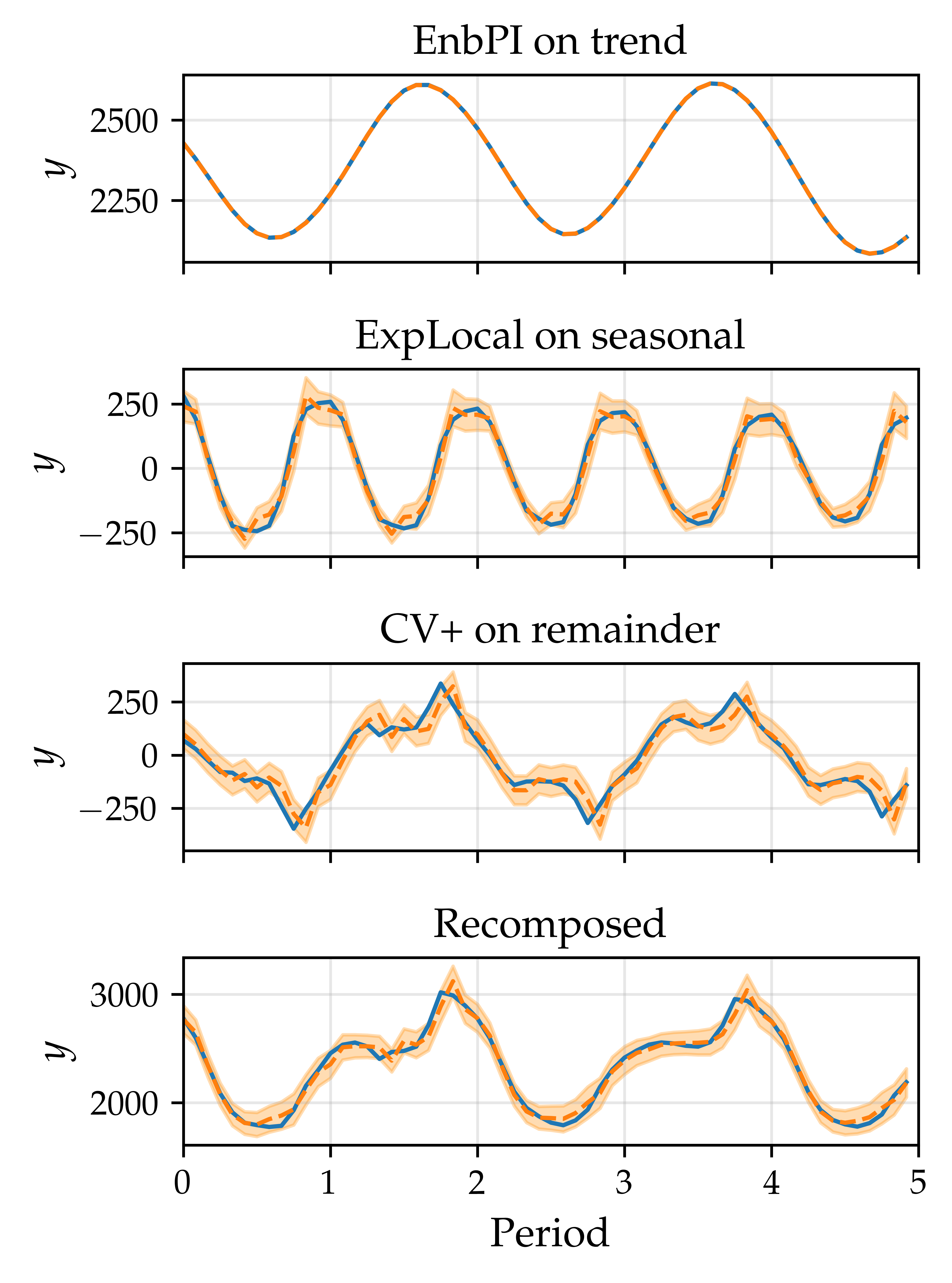}
    \caption{Time series and conformal prediction intervals per component and for the recomposition on \textbf{San Diego energy consumption} data. \emph{From left to right}: The seasonal component leverages BinaryPoint, BinaryLocal or ExpLocal.}
    \label{fig:decomposition_energy}
\end{figure}

\begin{figure}[H]
    \includegraphics[width=0.32\textwidth]{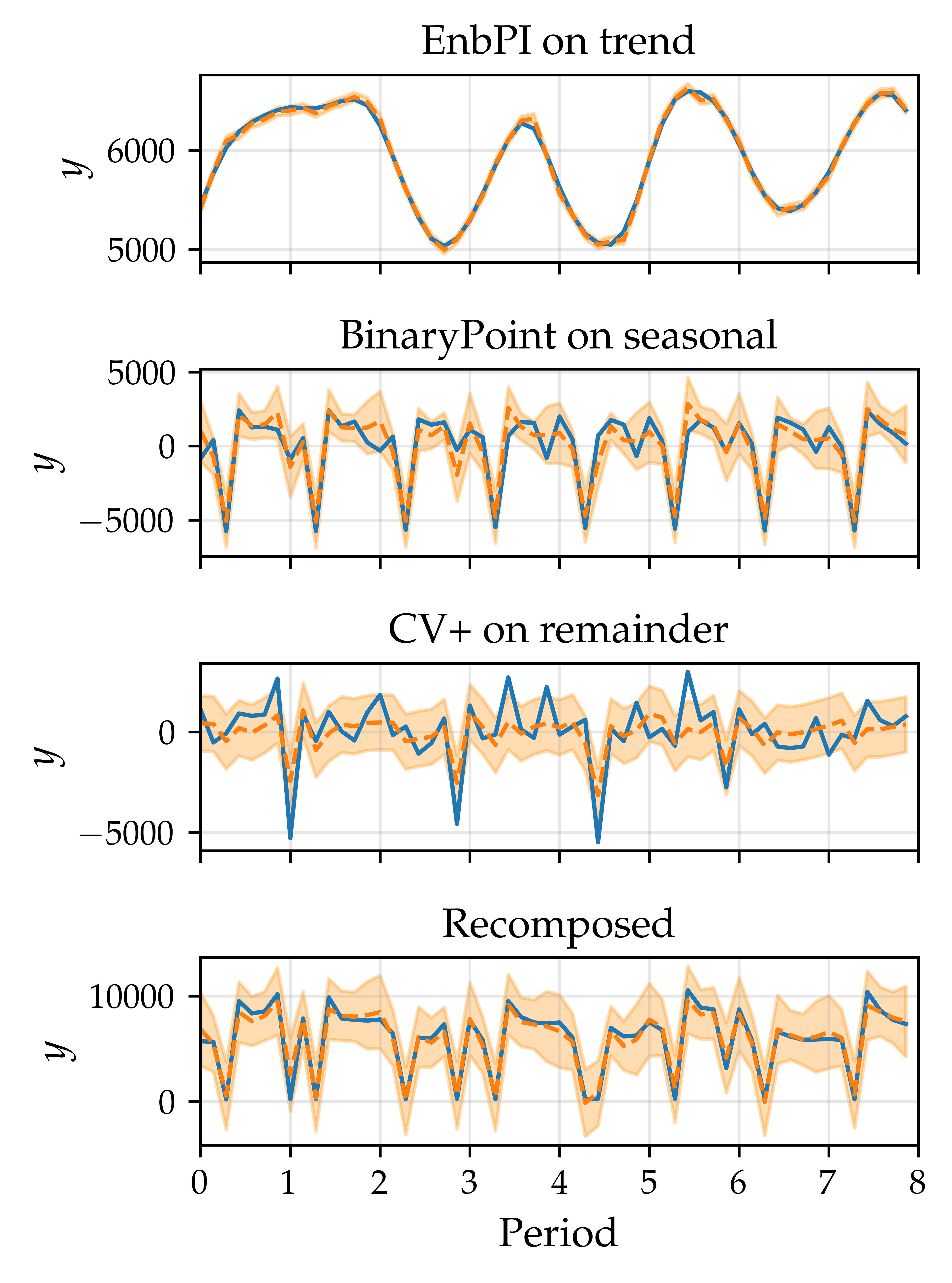}
    \includegraphics[width=0.32\textwidth]{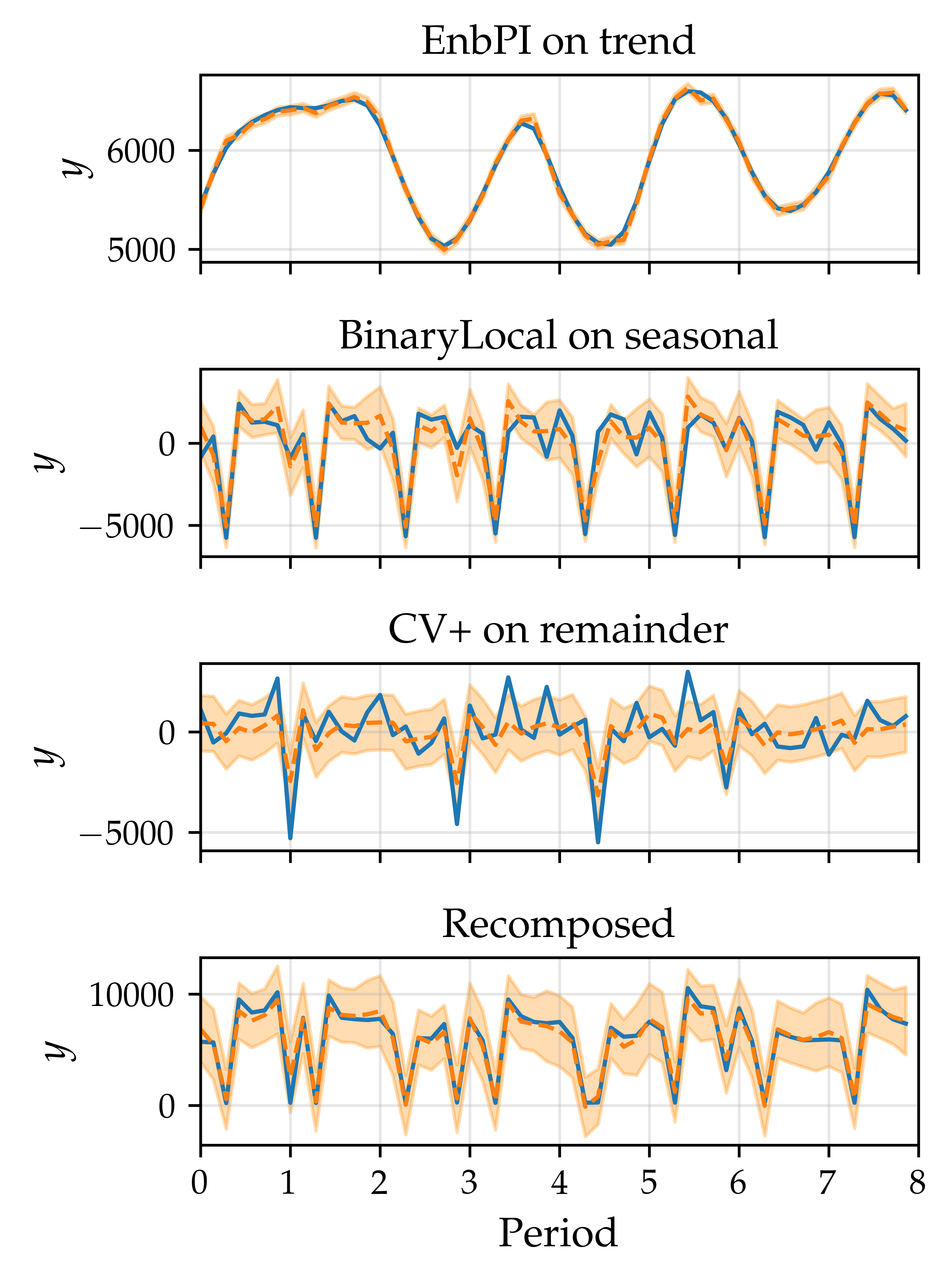}
    \includegraphics[width=0.32\textwidth]{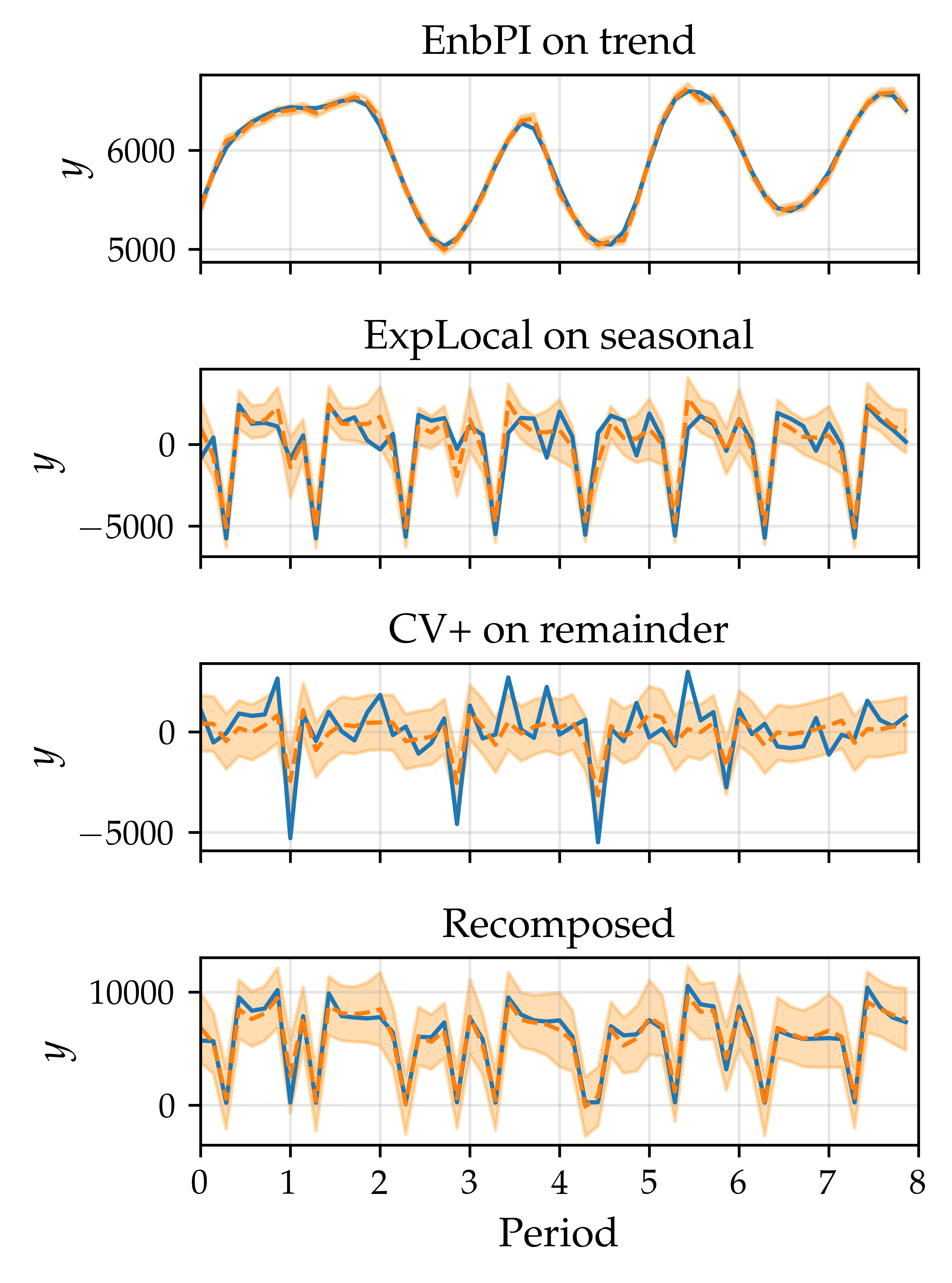}
    \caption{Time series and conformal prediction intervals per component and for the recomposition on \textbf{Rossmann store sales} data. \emph{From left to right}: The seasonal component leverages BinaryPoint, BinaryLocal or ExpLocal.}
    \label{fig:decomposition_sales}
\end{figure}

\begin{figure}[H]
    \includegraphics[width=0.32\textwidth]{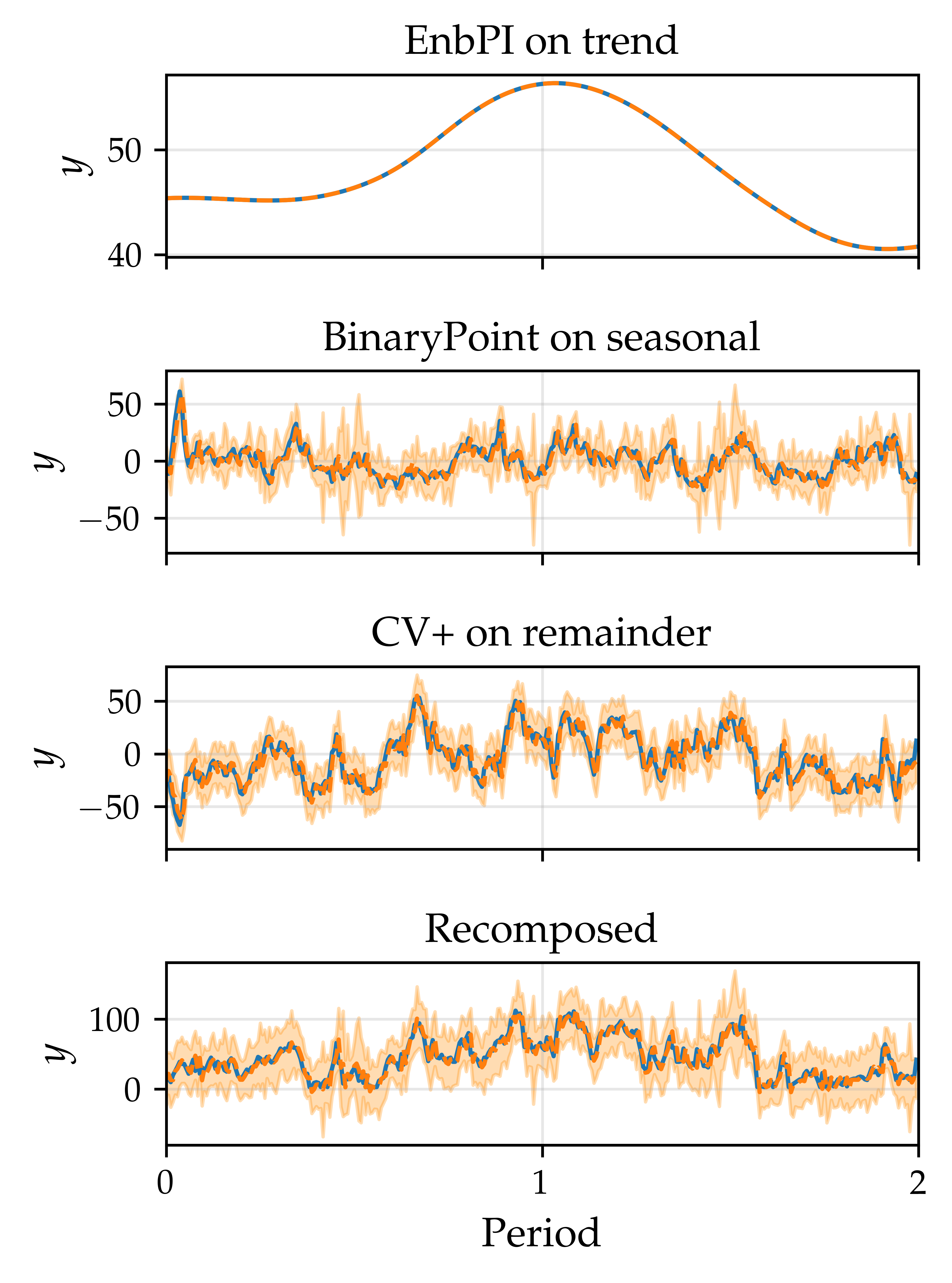}
    \includegraphics[width=0.32\textwidth]{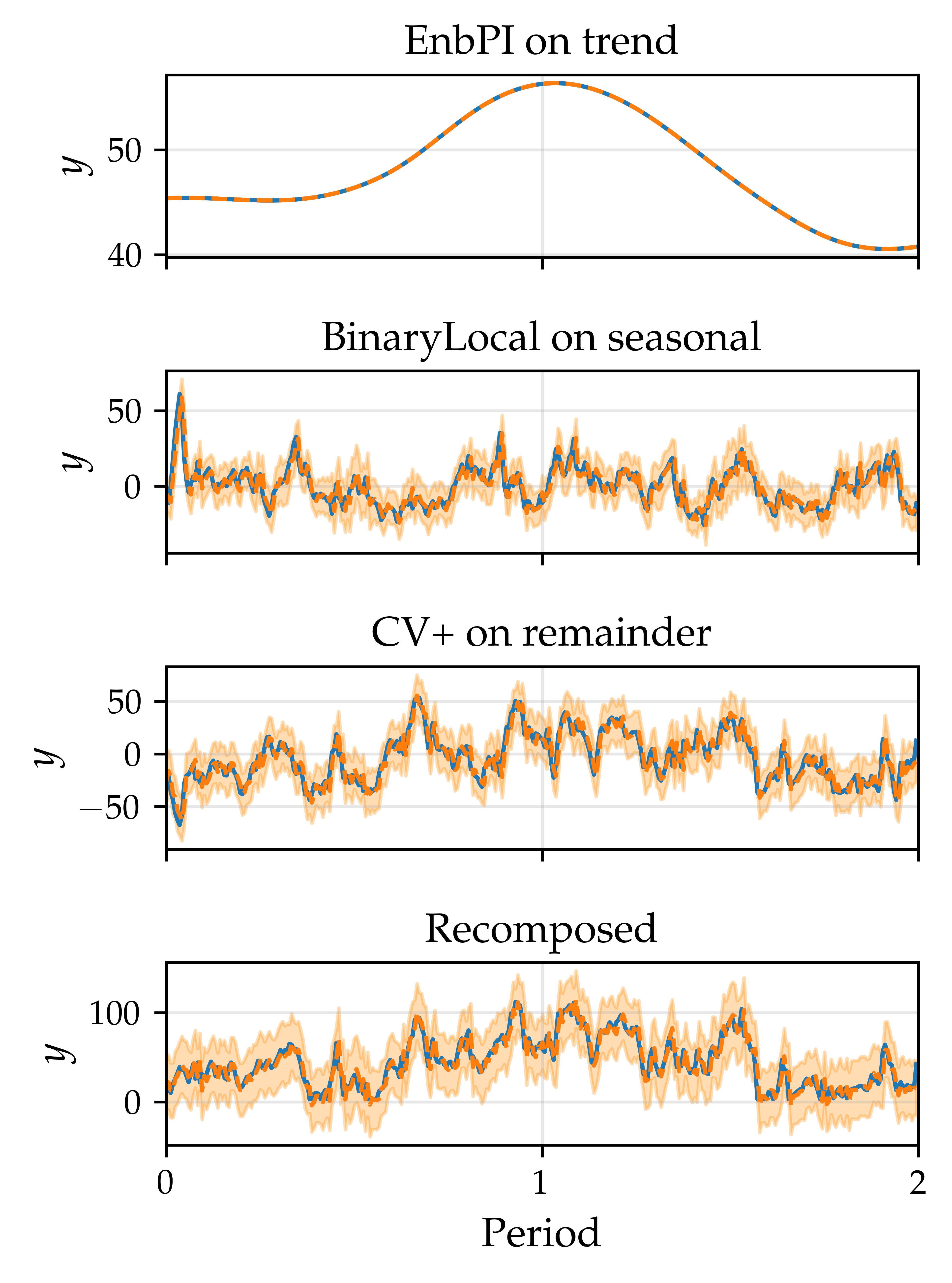}
    \includegraphics[width=0.32\textwidth]{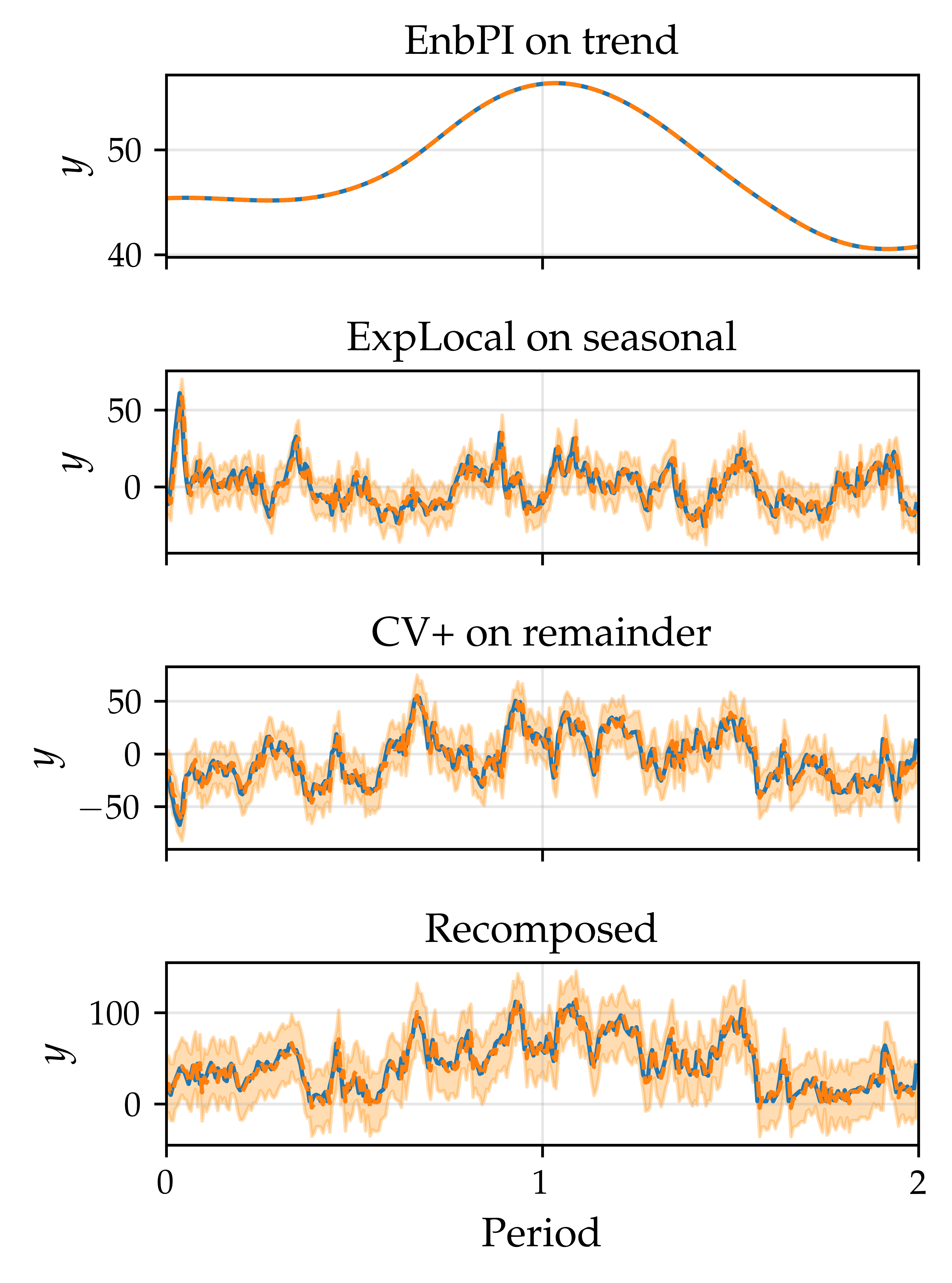}
    \caption{Time series and conformal prediction intervals per component and for the recomposition on \textbf{Beijing air quality} data. \emph{From left to right}: The seasonal component leverages BinaryPoint, BinaryLocal or ExpLocal.}
    \label{fig:decomposition_air_quality}
\end{figure}

Here, we display qualitative decomposition plots for every dataset: synthetic (\autoref{fig:decomposition_synthetic}), San Diego energy consumption (\autoref{fig:decomposition_energy}), Rossman store sales (\autoref{fig:decomposition_sales}) and Beijing air quality (\autoref{fig:decomposition_air_quality}). Specifically, we overlay an exemplary section of the original time series (\emph{blue}) with the conformal prediction intervals produced via our TSD approaches for every component (trend, season and remainder) as well as for the recomposition (\autoref{eq:recomp}). We display these plots three times, once for every seasonal weighting scheme (BinaryPoint, BinaryLocal, ExpLocal; see \autoref{subsec:season}). EnbPI and CV+ are used for the trend and remainder terms, respectively. We add some comments and observations below.

\paragraph{Synthetic data.} Similarly to \autoref{fig:synth} we observe tight PIs for the seasonal component using BinaryPoint. Conversely, wider intervals are observed for BinaryLocal and ExpLocal. The linear regressor correctly predicts zero mean for the $\mathcal{N}(0, 1)$ remainder component, producing a constant interval. The seasonal component clearly dominates the pattern of the time series, as also visible in the recomposition.

\paragraph{San Diego energy consumption.} A somewhat periodic trend is revealed, hinting at potentially multiple seasonalities of different time lengths, which could be investigated for example with MSTL \citep{bandara2021mstl} (but not the employed STL). For the seasonal patterns, wider intervals are observed at the extrema, which aim to compensate for the larger irregularities.

\paragraph{Rossmann store sales.} We observe complex trend and seasonal patterns which are associated with larger errors. The remainder component exhibits significant drops on certain days due to holidays without sales. As such, the remainder displays substantial volatility. Since the regressor tends to underpredict in these instances, it results in the generation of notably wide intervals for this component, contributing to the wide intervals and empirical overcoverage observed in the recomposition results shown in \autoref{tab:real_samp}.

\paragraph{Beijing air quality.} Here, we observe a notable difference between BinaryPoint and the other methods, BinaryLocal and ExpLocal, where the PIs of the former are characterized by spikes. It is hard to identify a consistent seasonal pattern, and season and remainder terms exhibit large similarity, suggesting that improvements on the decomposition quality can be made.

\newpage

\newpage
\section{Details on regression models.}
\label{app:rmse}

\paragraph{Predictive performance.} We evaluate the three underlying regressors (Linear, MLP and Gradient Boosting) via the \emph{root mean squared error} (RMSE), and report values in \autoref{tab:rmse_all_datasets}. Overall the trend is generally easy to predict, while the more irregular seasonal and remainder terms record higher errors. The linear and MLP regressors display similar results across datasets except on the sales data, where the MLP outperforms the linear model. Interestingly, this performance gap does not translate into lower PIAW scores when evaluating CP methods (see \autoref{app:res-synth}). Gradient boosting outperforms the other regressors on synthetic data due to its small sequential differences, but performs notably worse on the sales data, where abrupt changes in sales lead to large differences between time steps.

\paragraph{Parameter settings.} We also display hyperparameter settings for employed conformal algorithms and regression models in \autoref{tab:param-settings}. We do not perform extensive tuning and employ many default settings. The decay rate parameter $\lambda \in (0,1)$ for ExpLocal depends on each dataset since the period length varies across datasets. It is selected such that weights approach zero for data points that are separated by a full period length.

\captionsetup[table]{font=small}

\begin{table}[H]
    \centering
    \begin{minipage}[t]{0.45\textwidth}
        \centering
        \scalebox{0.7}{
        \begin{tabular}{lccc}
        \toprule\toprule
        \textbf{Component} & \textbf{Linear Reg.} & \textbf{MLP} & \textbf{Gradient Boost.} \\
        \midrule
        \multicolumn{4}{c}{Synthetic data} \\
        \midrule
        Original & 14.903 & 14.898 & 3.855 \\
        Trend & 0.0 & 0.002 & 0.003 \\
        Seasonal & 14.702 & 14.703 & 3.129 \\
        Noise & 1.001 & 1.003 & 1.242 \\
        \midrule
        \multicolumn{4}{c}{San Diego energy consumption} \\
        \midrule
        Original & 63.207 & 51.818 & 60.746 \\
        Trend & 1.336 & 3.443 & 2.552 \\
        Seasonal & 28.252 & 23.098 & 34.983 \\
        Noise & 38.903 & 32.755 & 45.179 \\
        \midrule
        \multicolumn{4}{c}{Rossmann store sales} \\
        \midrule
        Original & 539.921 & 359.879 & 1512.717 \\
        Trend & 28.003 & 66.599 & 41.372 \\
        Seasonal & 782.387 & 465.386 & 983.724 \\
        Noise & 897.016 & 758.889 & 1204.878 \\
        \midrule
        \multicolumn{4}{c}{Beijing air quality} \\
        \midrule
        Original & 16.86 & 17.226 & 18.296 \\
        Trend & 0.001 & 0.877 & 0.046 \\
        Seasonal & 8.735 & 8.753 & 8.994 \\
        Noise & 14.204 & 14.154 & 15.001 \\
        \bottomrule\bottomrule
        \end{tabular}
        }
        \caption{RMSE for various datasets across different components and models.}
        \label{tab:rmse_all_datasets}
    \end{minipage}
    \hspace{0.05\textwidth}
    \begin{minipage}[t]{0.45\textwidth}
        \centering
        \scalebox{0.7}{
            \begin{tabular}{lll}
            \toprule\toprule
            \textbf{CP algo.} & \textbf{Param.} & \textbf{Value} \\
            \midrule
            \multirow{4}{*}{EnbPI}    
            & \# bootstrap samples & 20 \\
            & window length & 1 \\ 
            & window overlap & False \\ 
            & seed & 42 \\ 
            \midrule
            \multirow{1}{*}{ACI} & $\gamma$ & 0.01 \\
            \midrule
            \multirow{1}{*}{CV+} & $k$ & 20 \\
            \midrule
            \multirow{2}{*}{BinaryPoint}    
            & use region & False \\
            & $\lambda$ & 1 \\ 
            \midrule
            \multirow{2}{*}{BinaryLocal}    
            & use region & True \\
            & $\lambda$ & 1 \\ 
            \midrule
            \multirow{2}{*}{ExpLocal}    
            & use region & True \\
            & $\lambda$ & $\in (0, 1)$ \\
            \midrule\midrule
            \textbf{Regressor} & \textbf{Param.} & \textbf{Value} \\
            \midrule
            \multirow{1}{*}{Linear Reg.}  
            & -- & -- \\
            \midrule
            \multirow{7}{*}{MLP}  
            & hidden layers & $(100, 100)$ \\
            & activation & ReLU\\
            & solver & Adam \\
            & LR init & $0.001$\\
            & LR & constant \\
            & max iter & 100 \\
            & early stop & True \\
            \midrule
            \multirow{3}{*}{Gradient Boost.}      
            & Mean loss & MSE \\
            & Quantile loss & Pinball \\
            & \# estimators & 100 \\
            \bottomrule\bottomrule
            \end{tabular}
        }
        \caption{Hyperparameter settings for different conformal algorithms and regressors.}
        \label{tab:param-settings}
    \end{minipage}
\end{table}

\end{document}